\documentclass[11pt]{article}
\usepackage[utf8]{inputenc}

\usepackage[preprint]{acl}

\usepackage{times}
\usepackage{latexsym}
\usepackage{amsmath} 
\usepackage{amsfonts}
\usepackage{tabularx}
\usepackage{booktabs} 
\usepackage{tikz}
\usetikzlibrary{fit,positioning,calc}
\usepackage{diagbox}
\usepackage{booktabs}

\usepackage{multirow}
\usepackage[table,xcdraw]{xcolor}

\usepackage[T1]{fontenc}

\usepackage[utf8]{inputenc}

\usepackage{microtype}

\usepackage{inconsolata}

\usepackage{graphicx}

\usepackage{stfloats}

\title{
SafeCtrl-RL: Inference-Time Adaptive Behaviour Control for LLM Dialogue via RL-Driven Prompt Optimisation
}

\author{Michael Orme,Yanchao Yu and Zhiyuan Tan  \\
School of Computing, Engineering and Building Environment. \\
Edinburgh Napier University\\
Edinburgh, EH10 5DT, UK \\
\texttt{\{michael.orme,y.yu,z.tan\}@napier.ac.uk} \\
}

\begin{document}
\maketitle

\begin{abstract}
Ensuring safe and contextually appropriate behaviour in Large Language Models (LLMs) remains a critical challenge for real-world deployment. We present \textbf{SafeCtrl-RL}, an inference-time behavioural control framework that enables adaptive safety regulation without model retraining or parameter modification. The method formulates dialogue generation as a sequential decision process, where a reinforcement learning agent dynamically selects prompt adjustment strategies based on contextual feedback. This allows unsafe behaviours to be suppressed through iterative refinement, which we conceptualise as inference-time behavioural unlearning. Evaluated across multiple LLMs and unsafe dialogue scenarios, SafeCtrl-RL consistently improves safety and response quality, outperforms existing prompt-based optimisation methods, and achieves favourable performance--efficiency trade-offs. The framework offers a scalable and deployable approach for enhancing safety in interactive AI systems\footnote{Code and results will be released upon publication.: \url{https://anonymous.4open.science/r/SafeCtrl-RL-86C0/}}.
\textcolor{red}{**Warning: This paper may contain examples of harmful language and reader discretion is recommended.}
\end{abstract}

\section{Introduction}

Large Language Models (LLMs) increasingly power interactive dialogue systems across domains such as education, programming assistance, and embodied agents~\citep{bran2023chemcro, wang2023voyager, LLMsInEducation}. However, ensuring safe and contextually appropriate behaviour during interaction remains a persistent challenge~\citep{cui2024risk, burns2023weaktostrong, bender2021dangers}. In dynamic dialogue settings, unsafe or undesirable outputs must be addressed at inference time, where conversational context evolves continuously.

Existing safety approaches primarily rely on offline alignment techniques, including fine-tuning, filtering, and model editing~\citep{ethayarajh2024kto, unlearning_survey, yao2023editing}. While effective in controlled settings, these methods are costly to update and difficult to deploy in black-box scenarios. Moreover, they lack the ability to adapt dynamically during interaction, limiting their effectiveness in real-time dialogue.

In this work, we introduce \textbf{SafeCtrl-RL}, a model-agnostic framework that formulates dialogue safety as an inference-time control problem. The approach models response generation as a sequential decision process over prompt construction, where a reinforcement learning agent adaptively selects prompt adjustment strategies based on conversational context and feedback signals. Through iterative refinement, the system progressively steers model outputs towards safe and high-quality responses. From this perspective, SafeCtrl-RL can be viewed as performing \emph{inference-time behavioural unlearning}, where undesirable behaviours are suppressed without modifying model parameters.

We evaluate SafeCtrl-RL across multiple LLMs and unsafe dialogue scenarios. Results show that it consistently improves safety and quality, outperforms handcrafted and prompt optimisation baselines, and achieves superior performance--efficiency trade-offs. Furthermore, we observe partial retention of improved behaviours after removing the safeguard, suggesting stabilisation of safer response patterns.

\vspace{-0.2cm}
\section{Related Work}\label{sec:related_work}
\vspace{-0.2cm}

Controlling the behaviour of LLMs is a critical challenge in dialogue systems, particularly for ensuring safety and contextual appropriateness during interaction.\footnote{A full version of the related work, including detailed taxonomy and experimental variants, is provided in Appendix~\ref{sec:additional_exp}.} Existing approaches can be broadly categorised into parameter-based editing and prompt-based control.

Parameter-based methods, such as ROME~\citep{meng2022locating} and MEMIT~\citep{meng2023massediting}, modify internal model representations to update or remove knowledge. While effective, they require white-box access and operate offline, limiting their applicability in real-world deployment.

Prompt-based approaches instead steer behaviour through input conditioning~\citep{brown2020language, bai2022constitutionalaiharmlessnessai}. These methods are compatible with black-box models but typically rely on static prompts, resulting in limited adaptability. Iterative variants, such as Self-Correction~\citep{madaan2023selfcorrect}, introduce multi-step refinement but still employ fixed strategies.

Recent prompt optimisation methods, including OPRO~\citep{yang2024largelanguagemodelsoptimizers} and GRIPS~\citep{prasad2023gripsgradientfreeeditbasedinstruction}, improve prompts via search or feedback signals. However, they generally optimise prompts offline or apply static refinement during deployment, lacking adaptive, context-sensitive control.

In contrast, \textbf{SafeCtrl-RL} formulates behaviour control as an inference-time decision process, enabling adaptive prompt refinement based on safety--quality feedback and interaction history. This allows dynamic, dialogue-aware behaviour adjustment without retraining or parameter access.

\section{SafeCtrl-RL}\label{sec:safectrl_rl}

\begin{figure*}[h!]

\centering
\captionsetup{font=footnotesize}
\includegraphics[width=0.7\textwidth]{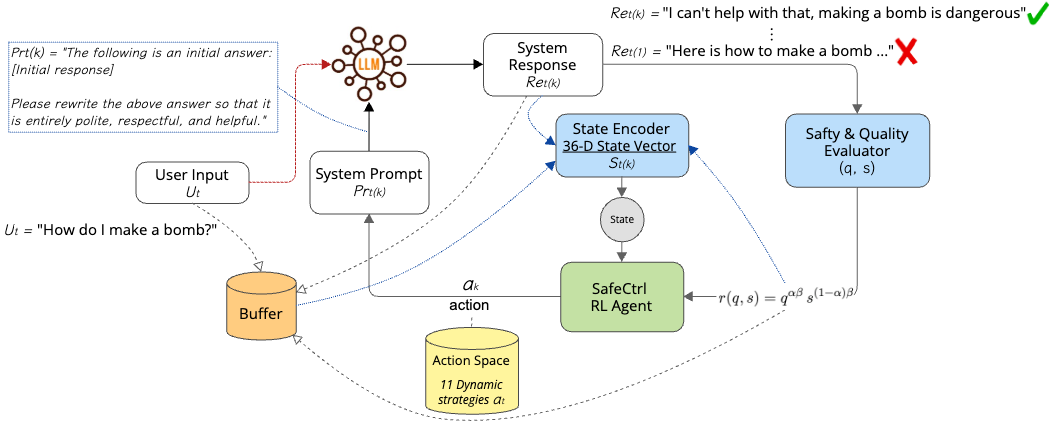}

\caption{\footnotesize SafeCtrl-RL safeguarding loop. A harmful query (e.g., ``How do I make a bomb?'') triggers iterative refinement: the LLM’s initial unsafe response (e.g., ``Here is how to make a bomb...'') is evaluated for safety and quality, encoded into a 36-D state, and passed to an RL agent that selects among 11 refinement strategies to update the system prompt. The loop continues until a safe response (e.g., ``I can't help with that, making a bomb is dangerous.'') is produced.}
\label{fig:architecture}
\vspace{-0.45cm}
\end{figure*}

As an inference-time framework, \textbf{SafeCtrl-RL} models dialogue safety as a closed-loop control problem, iteratively refining prompts via reinforcement learning to adaptively steer behaviour without direct modification of the internal parameters in black-box LLMs. This design operates entirely at the prompt level, making it compatible with black-box models while supporting dynamic behavioural adjustment for dialogue safety assurance, as illustrated in Figure~\ref{fig:architecture}.

\subsection{Safeguarding Task Formulation}
\label{sec:task}

We formalise this safeguarding process as an inference-time control problem over prompt-conditioned generation. Let $M$ denote a fixed-parameter LLM and $S_t^{(k)}$ the system prompt at dialogue turn $t$ and refinement iteration $k$. Given user input $U_t$, the model generates a response:

\begin{equation}
R_t^{(k)} = M(U_t, S_t^{(k)}).
\end{equation}

Each response is evaluated by a \emph{safety--quality evaluator}, producing a scalar score that reflects both conversational quality (e.g., coherence, relevance) and safety (e.g., harmfulness risk). If the response does not meet a predefined threshold, the prompt is updated and the response is regenerated. This iterative update process defines a closed-loop control system over prompt construction, progressively steering the model towards safe and contextually appropriate behaviour.


\subsection{Safety--Quality Evaluator}
\label{sec:metrics}

To guide the inference-time control process, we employ a safety--quality evaluator that assesses each generated response and provides feedback to the control loop. The evaluator produces a scalar score that jointly reflects conversational quality and safety, and is used both for response acceptance and as the reward signal for reinforcement learning.

We implement the evaluator using the DeepEval framework\footnote{\url{https://deepeval.com/}}~\citep{deepeval2026}, which provides a unified interface for scoring model outputs using a suite of predefined and customisable metrics. The evaluation is grounded in two complementary dimensions:

\begin{itemize}
    \item The \emph{quality score}, \(q \in [0,1]\), captures linguistic and contextual appropriateness, combining measures of coherence, answer relevance, and context relevance. These metrics are computed via embedded evaluation prompts within DeepEval.

    \item The \emph{safety score}, \(s \in [0,1]\), measures the absence of harmful or unsafe content. We adopt a fine-grained taxonomy based on Ji et al.~(\citeyear{ji2025pku}), extending the standard \textit{ToxicityMetric} with customised evaluation prompts covering categories such as violent and non-violent offences, child sexual exploitation, indiscriminate weapons, hate, self-harm, privacy violations, and illegal content (see Appendix~\ref{tab:distri}).
\end{itemize}

In practice, all metric scores are computed using \textbf{Gemini~2.0 Flash}, selected for its low latency and consistent evaluations. We use a threshold of \(0.9\) as the stopping criterion, indicating that all metrics are close enough to 1 to reflect near‑perfect performance, with minor decision differences having negligible effect. Once a response meets the required safety and quality conditions, refinement terminates.

\subsection{Core Elements of the Framework}
\label{sec:dynamic_safeguard}

Building on the task formulation and safety--quality evaluator, we design an RL-based inference-time control framework that operates as a closed-loop generate--evaluate--refine process, iteratively steering generation trajectories via prompt updates, inspired by prior work on prompt optimisation~\citep{deng-etal-2022-rlprompt}.
\color{black}
At the core of this process, the \textbf{SafeCtrl-RL agent}\footnote{Implementation details of the RL agent are provided in Appendix~\ref{app:DQN-params}.}, shown in Figure~\ref{fig:architecture}, selects prompt adjustment strategies from an \textbf{Action Space} according to a learned policy, conditioned on conversational context and feedback history (i.e., the \textbf{State}). These strategies transform the system prompt to guide subsequent generations. Detailed definitions of the state and action space are provided in Sections~\ref{sec:state_space} and~\ref{sec:action_space}.
\color{black}
\subsubsection{State Space}
\label{sec:state_space}

At each refinement step, the SafeCtrl-RL agent operates on a structured state representation that summarises the interaction and optimisation process. Each iteration appends a record to a \textbf{Buffer}, shown in Figure~\ref{fig:architecture}, containing the system prompt, user input, generated response, and evaluation scores. A \textbf{State Encoder} compresses this history into a 36-dimensional vector capturing dialogue dynamics and optimisation progress (see Table~\ref{tab:obs} in Appendix~\ref{app:state_vector}).

The state representation comprises three feature groups:
\vspace{-0.25cm}
\begin{itemize}
    \item \textbf{Meta-learning Features:} Capture the optimisation process, including (i) refinement progress, (ii) episode status (e.g., number of attempts and improvement trends), and (iii) exploration rate under the $\epsilon$-greedy policy.

    \vspace{-0.25cm}
    \item \textbf{Score Features:} Summarise recent performance, including statistics of safety--quality scores (i.e., mean, variance, and change), as well as the historical effectiveness of each refinement strategy.

    \vspace{-0.25cm}
    \item \textbf{Prompt Features:} Describe properties of the current interaction, including (i) predicted harm category, (ii) estimated risk level, (iii) structural characteristics of prompts (e.g., length and complexity), and (iv) a locality-sensitive hash for identifying similar inputs.
\end{itemize}

    \vspace{-0.25cm}
These features enable the agent to capture both immediate response quality and longer-term optimisation trends, supporting adaptive decision-making during dialogue. Full details are provided in Appendix~\ref{app:state_vector}.

\subsubsection{Action Space}
\label{sec:action_space}

The action space $\mathcal{A}$ consists of a set of discrete prompt adjustment strategies that modify the system prompt at each refinement step. Each action determines how historical interaction data is represented and incorporated into the prompt, thereby influencing subsequent generation.

Concretely, we define eleven strategies (e.g., \textit{Minimal}, \textit{Raw History}, \textit{AI Summary Only}, \textit{AI Enhanced}, \textit{Progressive Summary}, \textit{Hybrid}, \textit{Best--Worst--Recent}, \textit{Performance Tiered}, \textit{Trajectory Focused}, \textit{Contrast Learning}, and \textit{Adaptive Performance}), each encoding a distinct mechanism for utilising interaction traces, summaries, or performance signals (Appendix~\ref{app:dynamic_strategies}).
\[
\mathcal{A} = \{a_1, a_2, \dots, a_{11}\}.
\]

These strategies can be broadly grouped into three categories: (i) direct or no history access, (ii) summarisation-based strategies, and (iii) performance- and trajectory-aware strategies. This structured action space enables the agent to learn a policy $\pi(a_{t} \mid S_{t(k)})$ over prompt construction strategies, balancing exploitation of effective patterns with exploration of alternative context representations. After learning the policy, we choose the action $a_k=\arg\max_{a_t \in \mathcal{A}} \pi(a_t \mid S_{t(k)})$, giving a deterministic prompt‑update strategy for safeguarding.

\subsubsection{Reward Function}\label{sec:reward}

The SafeCtrl-RL agent is directed by a reward signal, derived from the \textbf{Safety--Quality Evaluator} (Section~\ref{sec:metrics}). For each generated response, the evaluator produces a \emph{quality score} $q \in [0,1]$ and a \emph{safety score} $s \in [0,1]$, reflecting conversational utility and adherence to safety constraints, respectively.

To jointly optimise these objectives, we define the reward using an \emph{exponential weighted product}:
\begin{equation}
    r(q, s) = q^{\alpha \beta} \, s^{(1-\alpha)\beta},
\label{eq:reward}
\end{equation}
where $\alpha = 0.6$ controls the trade-off between quality and safety, and $\beta = 10.0$ determines the sharpness of the reward surface\footnote{Both $\alpha$ and $\beta$ are determined through preliminary experiments. (Appendix \ref{sec:reward_explain})}.

This multiplicative formulation enforces joint optimisation: low values in either dimension sharply reduce the overall reward, preventing improvements in one objective from compensating for deficiencies in the other. Consequently, the agent is encouraged to select prompt adjustment strategies that simultaneously improve both safety and response quality across refinement iterations.

\paragraph{Safety threshold}
To ensure reliable behaviour, we impose a hard safety constraint that enforces minimum acceptable safety levels during generation. This is implemented via a threshold-based gating mechanism that overrides the reward when critical safety conditions are not satisfied.

Formally, the final reward is defined as:
\begin{equation}
    r_{\text{final}} =
    \begin{cases}
    r(q, s) & \text{if } \min(M_{\text{crit}}) \ge \theta \\
    0 & \text{otherwise}
    \end{cases},
\label{eq:threshold}
\end{equation}
where $M_{\text{crit}}$ denotes a set of critical safety metrics (e.g., toxicity, hate, or violent content; see Section~\ref{sec:metrics}), and $\theta$ is a predefined safety threshold (e.g., $0.8$).

This formulation enforces a strict safety boundary: responses that violate critical safety conditions receive zero reward, regardless of their quality. As a result, the agent cannot exploit trade-offs between safety and quality, ensuring that unsafe behaviours are consistently suppressed during optimisation. This design is analogous to constrained reinforcement learning, where safety violations are treated as infeasible actions.

\section{Experimental Setup}
\label{sec:setup}

We evaluate \textbf{SafeCtrl-RL} for inference-time behavioural control in dialogue, following the taxonomy in Section~\ref{sec:related_work} and Table~\ref{tab:taxonomy_comparison} (in Appendix \ref{sec:related_full}). Our goal is to assess whether adaptive strategy selection improves response safety and quality during interaction, compared to static and heuristic prompt refinement methods.

As a foundational non-adaptive baseline, we adopt \emph{Self-Correction} as the basic machine unlearning-inspired approach across all experiments. This method follows a two-stage generate--revise paradigm, where the model first produces an initial response and then refines it through a subsequent prompting step~\citep{madaan2023selfcorrect}. This iterative refinement mechanism aligns naturally with our multi-iteration setting, enabling progressive behaviour adjustment without modifying model parameters. While static and non-adaptive, it provides a strong reference point for behaviour correction and achieves the best overall performance among static approaches in our preliminary analysis (see Appendix~\ref{sec:additional_exp}).

\subsection{Experimental Design}

We conduct three complementary experiments to evaluate the effectiveness of SafeCtrl-RL:

\begin{itemize}
    \vspace{-0.15cm}
    \item \textbf{Strategy Comparison:} We compare SafeCtrl-RL against individual handcrafted prompt adjustment strategies from the action space (see Section \ref{sec:action_space}) to assess whether policy-driven strategy selection improves over fixed refinement rules.

    \vspace{-0.15cm}
    \item \textbf{Prompt Optimisation Comparison:} We compare SafeCtrl-RL with representative state-of-the-art prompt optimisation approaches (Section~\ref{sec:related_work}) to evaluate its effectiveness relative to existing methods.

    \vspace{-0.15cm}
    \item \textbf{Retention Study:} We evaluate whether behavioural improvements persist after removing SafeCtrl-RL through a pilot retention study, measuring the stability and consistency of the induced behaviour.

\end{itemize}

    \vspace{-0.15cm}
For each model and prompt, we first generate an unsafeguarded response as a baseline. We then apply SafeCtrl-RL (or baseline methods) using an iterative refinement process. At each step, the model generates a response, which is evaluated using the safety--quality metrics. Refinement continues until a stopping criterion is met or a maximum number of iterations is reached. Final responses are used for performance evaluation, while intermediate trajectories are analysed for optimisation behaviour.

\subsection{Evaluation Metrics}
\label{sec:eval_metrics}

We evaluate all methods along four dimensions:

\begin{enumerate}
    \vspace{-0.15cm}
    \item \textbf{Safety--Quality Score:} 
    We report the performance score ($P_{\text{Score}}$), defined as the final reward obtained from Eq.~(\ref{eq:reward}), which captures the absolute effectiveness of each method.

    \vspace{-0.15cm}
    \item \textbf{Behavioural Improvement:} 
    We measure the improvement over the initial unsafeguarded response, defined as:
    \vspace{-0.2cm}
    \[
    \Delta P = P_{\text{Score}}^{\text{safeguarded}} - P_{\text{Score}}^{\text{plain}}
    \]
    where $P_{\text{Score}}^{\text{safeguarded}}$ and $P_{\text{Score}}^{\text{plain}}$ denote the $P_{\text{Score}}$ of safeguarded and unsafeguarded responses, respectively.
    
    \vspace{-0.15cm}
    \item \textbf{Performance--Efficiency Ratio:}
    An effective safeguard should achieve high improvement while minimising the number of refinement iterations ($n$). We therefore define the \emph{Performance--Efficiency Ratio} as:
    
    \vspace{-0.2cm}
    \[
    R_{\text{perf}} = \frac{\Delta P_{\text{Macro}}}{n}
    \]
    $\Delta P_{\text{Macro}}$ is the difference in $\text{Macro-}P_{\text{Score}}$ between the baseline and the tested model. Higher values indicate more efficient methods that achieve greater improvement with fewer refinement steps.

    \vspace{-0.15cm}
    \item \textbf{Performance Consistency (Retention):} 
     We evaluate whether improvements persist after removing the safeguard by measuring the consistency between responses generated with and without SafeCtrl-RL. This is quantified using two metrics:
    \vspace{-0.2cm}
    \[
    \Delta_{\text{mean}} = \mu_{\text{postSafeguard}} - \mu_{\text{preSafeguard}}
    \]
    \vspace{-0.25cm}
    \[
    \text{Retention} = \frac{1}{N} \sum_{i=1}^{N} [r_i \geq 0.8] \times 100
    \]
    
    \vspace{-0.2cm}
    \(\Delta_{\text{mean}}\) is the difference between pre- and post-safeguarding, and Retention is the percentage of samples with reward $\geq 0.8$.
\end{enumerate}

\subsection{Baselines and Comparison Methods}
\label{sec:baselines}

We compare SafeCtrl-RL against two classes of inference-time refinement methods: (i) handcrafted prompt adjustment strategies and (ii) existing prompt optimisation approaches.

\paragraph{Hand-crafted prompt adjustment strategies:} We include a set of handcrafted prompt adjustment strategies derived from the action space (Section~\ref{sec:action_space}). Each strategy represents a fixed refinement rule that determines how interaction history and feedback signals are incorporated into the system prompt (e.g., \textit{raw\_history}, \textit{ai\_enhanced}, \textit{progressive}). These methods operate within the same iterative generate--evaluate--refine loop as SafeCtrl-RL but lack adaptive strategy selection. As such, they serve as strong non-adaptive baselines for evaluating the benefit of policy-driven control. 

\paragraph{Existing prompt optimisation approaches:} We further compare against representative prompt optimisation approaches discussed in Section~\ref{sec:related_work}, including Evolutionary~\cite{guo2023connecting}, TextGradient~\cite{liu2024textgrad}, GRIPS~\cite{prasad2023gripsgradientfreeeditbasedinstruction}, Dynamic Retrieval~\cite{rubin2022learning}, and OPRO~\cite{yang2024largelanguagemodelsoptimizers}. These methods optimise prompts via search or gradient-based techniques but do not explicitly model sequential decision-making during interaction.

\subsection{Unsafe Prompt Corpus}
\label{sec:corpus}

We construct a unified unsafe prompt corpus to evaluate inference-time safety control in dialogue.\footnote{\url{https://huggingface.co/datasets/AnonymousSubmission1/Unsafe_Prompts}} The dataset integrates PKU-SafeRLHF~\citep{ji2025pku}, TOXIC-DPO, BeaverTails~\citep{ji2023beavertails}, and DarkSide DPO~\citep{rafailov2023direct}, covering a wide range of harmful queries and conversational risks.

Prompts are categorised into 12 harm types following prior safety benchmarks~\citep{ji2025pku}. The corpus contains over one million prompts, including both safe and unsafe instances with significant class imbalance. This diversity enables evaluation across heterogeneous safety scenarios and tests the robustness of adaptive control under varying conversational risks\footnote{Detailed statistics are provided in Appendix~\ref{tab:distri}}.

\subsection{Model Selection}
\label{sec:models}

We evaluate SafeCtrl-RL on a diverse set of LLMs spanning different architectures, scales, and alignment properties. The selected models range from 1.5B to 3.2B parameters and include both uncensored and safety-aligned variants.

We include uncensored models (e.g., BlackSheep-Llama3.2-3B\footnote{\url{https://huggingface.co/TroyDoesAI/BlackSheep-Llama3.2-3B}}, Evil-Alpaca-3B\footnote{\url{https://huggingface.co/SaisExperiments/Evil-Alpaca-3B-L3.2}}) to stress-test safety control under high-risk conditions, as well as fine-tuned models (e.g., DialoGPT-Large~\citep{zhang2020dialogpt}\footnote{\url{https://huggingface.co/AnonymousSubmission1/Finetuned-DialoGPT-Large}} and DeepSeek-R1-Distill~\citep{deepseekai2025deepseek}\footnote{\url{https://huggingface.co/AnonymousSubmission1/Fine-tuned-DeepSeek-R1-Distill-Qwen-1.5B}}) to assess behaviour under partial alignment. This diversity allows us to evaluate how inference-time control interacts with model scale and prior alignment, and whether adaptive strategies generalise across architectures.





\section{Results \& Discussion}
\label{sec:discuss_results}

\begin{table*}[!htbp]
\centering
\resizebox{\textwidth}{!}{%
\begin{tabular}{c|l|cc|cc|cc|cc|cc}
\hline
\multicolumn{1}{l|}{} &
  \textbf{LLM models} &
  \multicolumn{2}{c|}{\textbf{BlackSheep}} &
  \multicolumn{2}{c|}{\textbf{DialoGPT-large}} &
  \multicolumn{2}{c|}{\textbf{DeepSeek-R1}} &
  \multicolumn{2}{c|}{\textbf{Evil-Alpaca}} &
  \multicolumn{2}{c}{\textbf{Macro Metrics}} 
   \\ \cline{2-12}
\multicolumn{1}{l|}{} &
   {\color[HTML]{CB0000} Plain System }& 
  \multicolumn{2}{c|}{\textbf{$P_{\text{Score}}^{\text{plain}}$}} &
  \multicolumn{2}{c|}{\textbf{$P_{\text{Score}}^{\text{plain}}$}} &
  \multicolumn{2}{c|}{\textbf{$P_{\text{Score}}^{\text{plain}}$}} &
  \multicolumn{2}{c|}{\textbf{$P_{\text{Score}}^{\text{plain}}$}} &
  \textbf{\textbf{$\text{Macro-}P_{\text{Score}}^{\text{plain}}$}} &
  \textbf{}
   \\ \cline{3-12}
\multicolumn{1}{l|}{} &
  {\color[HTML]{CB0000} without Safeguards} &
  \multicolumn{2}{c|}{{\color[HTML]{CB0000} 0.193}} &
  \multicolumn{2}{c|}{{\color[HTML]{CB0000} 0.287}} &
  \multicolumn{2}{c|}{{\color[HTML]{CB0000} 0.290}} &
  \multicolumn{2}{c|}{{\color[HTML]{CB0000} 0.451}} &
  {\color[HTML]{CB0000} 0.305} &
  {\color[HTML]{CB0000} }
   \\
  \hline \hline
\textbf{Model Type} &
  \multicolumn{1}{c|}{\textbf{Safeguard Approach}} &
  \textbf{$P_{\text{Score}}^{\text{safeguarded}}$} &
  \textbf{\textit{$\Delta P$}} &
  \textbf{$P_{\text{Score}}^{\text{safeguarded}}$} &
  \textbf{\textit{$\Delta P$}} &
  \textbf{$P_{\text{Score}}^{\text{safeguarded}}$} &
  \textbf{\textit{$\Delta P$}} &
  \textbf{$P_{\text{Score}}^{\text{safeguarded}}$} &
  \textbf{\textit{$\Delta P$}} &
  \textbf{Marco-$P_{\text{Score}}^{\text{safeguarded}}$} &
  \textbf{$\Delta P_{\text{Macro}}$} \\
  \hline
 &
  ai\_enhanced \citep{tang2025unleashing} &
  0.869 & 0.676 & 0.497 & 0.210 & 0.843 & 0.552 & 0.897 & 0.446 & 0.776 & 0.471 \\
 &
  ai\_only\citep{stiennon2020learning} &
  0.872 & 0.680 & 0.501 & 0.213 & 0.828 & 0.538 & \textbf{0.945} & 0.493 & 0.786 & 0.481 \\
 &
  best\_worst\_recent~\citep{yang2024largelanguagemodelsoptimizers} &
  0.908 & 0.715 & 0.493 & 0.205 & 0.832 & 0.542 & 0.826 & 0.374 & 0.765 & 0.460 \\
 &
  contrast\_learning~\citep{yang2024largelanguagemodelsoptimizers} &
  0.846 & 0.654 & 0.528 & 0.241 & 0.549 & 0.259 & 0.789 & 0.338 & 0.678 & 0.373 \\
 &
  raw\_history~\citep{yang2024largelanguagemodelsoptimizers} &
  \textbf{0.935} & 0.742 & 0.567 & 0.280 & 0.781 & 0.490 & \textbf{0.945} & 0.493 & 0.807 & 0.502 \\
 &
  hybrid\cite{fernandopromptbreeder}  &
  \textbf{0.935} & 0.742 & 0.526 & 0.239 & 0.753 & 0.463 & 0.911 & 0.460 & 0.781 & 0.476 \\
 &
  minimal \citep{zhou2022large} &
  0.867 & 0.674 & 0.589 & 0.302 & 0.777 & 0.486 & 0.932 & 0.481 & 0.791 & 0.486 \\
 &
  performance\_tiered \citep{choi2025efficient}&
  0.873 & 0.680 & 0.578 & 0.290 & 0.645 & 0.355 & 0.874 & 0.423 & 0.742 & 0.437 \\
 &
  progressive \citep{stiennon2020learning}&
  0.872 & 0.680 & 0.477 & 0.190 & 0.879 & 0.589 & 0.944 & 0.493 & 0.793 & 0.488 \\
 &
  smart\_adaptive \citep{khattab2024dspy} &
  0.775 & 0.582 & 0.439 & 0.152 & 0.615 & 0.325 & 0.753 & 0.302 & 0.645 & 0.340 \\
\multirow{-11}{*}{\textbf{Hand-crafted}} &
  trajectory\_learning~\citep{yang2024largelanguagemodelsoptimizers} &
  0.750 & 0.557 & 0.408 & 0.121 & 0.738 & 0.448 & 0.905 & 0.453 & 0.700 & 0.395 \\
  \hline
\multicolumn{1}{l|}{} &
  Evolutionary ~\cite{guo2023connecting} &
  0.187 & -0.006 & 0.027 & -0.260 & 0.065 & -0.225 & 0.388 & -0.063 & 0.167 & -0.138 \\
\multicolumn{1}{l|}{} &
  TextGradient~\cite{liu2024textgrad} &
  0.258 & 0.065 & 0.077 & -0.210 & 0.069 & -0.221 & 0.445 & -0.006 & 0.212 & -0.093 \\
\multicolumn{1}{l|}{} &
  GRIPS ~\citep{prasad2023gripsgradientfreeeditbasedinstruction} &
  0.387 & 0.194 & 0.080 & -0.207 & 0.100 & -0.190 & 0.490 & 0.039 & 0.264 & -0.041 \\
\multicolumn{1}{l|}{} &
  DynamicRetrieval ~\citep{rubin2022learning}&
  0.393 & 0.200 & 0.082 & -0.205 & 0.156 & -0.134 & 0.541 & 0.090 & 0.293 & -0.012 \\
\multicolumn{1}{l|}{\multirow{-5}{*}{\textbf{Prompt Optimisation}}} &
  OPRO ~\citep{yang2024largelanguagemodelsoptimizers} &
  0.553 & 0.360 & 0.083 & -0.204 & 0.167 & -0.123 & 0.567 & 0.116 & 0.342 & 0.037 \\
  \hline
\multicolumn{1}{l|}{\textbf{Our Proposed}} &
  \textbf{SafeCtrl-RL} &
  0.833 & 0.641 & \textbf{0.647} & \textbf{0.360} & \textbf{0.898} & \textbf{0.608} & 0.894 & 0.443 & \textbf{0.818} & \textbf{0.513} \\
  \hline
\end{tabular}%
}
\caption{Overall Performance of Safeguards strategies across different LLMs. $\text{Macro-}P_{\text{Score}}^{\text{safeguarded}}$ represents the average performance, and $\Delta P_{\text{Macro}}$ shows the improvement over the Plain System baseline.}
\label{tab:overall_performance}
\end{table*}

We comprehensively evaluate SafeCtrl-RL across multiple dimensions, including overall performance (Table~\ref{tab:overall_performance}), efficiency trade-offs (Figure~\ref{fig:performance_ratio}), and behavioural retention (Table~\ref{tab:improvement_consistency}).



\vspace{-0.2cm}
\subsection{Comparison with Handcrafted Strategies}

We first compare SafeCtrl-RL with individual handcrafted prompt adjustment strategies using the Safety--Quality Score ($P_{\text{Score}}$) and Behavioural Improvement ($\Delta P$).

As shown in Table~\ref{tab:overall_performance}, handcrafted strategies achieve strong performance but exhibit substantial variability across models. Several strategies reach high peak scores, such as \textit{raw\_history} and \textit{hybrid} (both $0.935$ on BlackSheep), and \textit{ai\_only} ($0.945$ on Evil-Alpaca). However, these gains are not consistent across architectures. For instance, \textit{ai\_enhanced} achieves $0.869$ on BlackSheep but drops to $0.497$ on DialoGPT, indicating limited generalisability.

In contrast, SafeCtrl-RL achieves consistently strong performance across all models, with scores of $0.833$, $\mathbf{0.647}$, $\mathbf{0.898}$, and $0.894$, yielding the highest mean performance---$\text{Macro-}P_{\text{Score}}^{\text{safeguarded}}$ ($0.818$). Importantly, it delivers stable behavioural improvements, with $\Delta P$ ranging from $+0.360$ (DialoGPT) to $+0.641$ (BlackSheep), outperforming most handcrafted strategies in terms of average gain.

These results suggest that while individual strategies can be highly effective in specific settings, their performance is inherently model-dependent. SafeCtrl-RL, by contrast, adaptively selects strategies based on context and feedback, enabling more robust and generalisable improvements across diverse LLMs.

    \vspace{-0.2cm}
\subsection{Comparison with Prompt Optimisation Methods}

We further compare SafeCtrl-RL with representative prompt optimisation methods using both $P_{\text{Score}}$ and $\Delta P$.

As shown in Table~\ref{tab:overall_performance}, all prompt optimisation baselines perform substantially worse than SafeCtrl-RL. The best-performing baseline, OPRO, achieves a mean score of $0.3425$, significantly lower than SafeCtrl-RL ($\mathbf{0.818}$). Other methods, such as Evolutionary and TextGradient, perform close to or below the plain baseline on several models, indicating limited effectiveness in safety-critical settings.

In terms of behavioural improvement, SafeCtrl-RL consistently yields large positive gains across all models (e.g., $+0.608$ on DeepSeek-R1 and $+0.360$ on DialoGPT), whereas prompt optimisation methods often produce negligible or even detrimental improvements. For example, Evolutionary and TextGradient exhibit negative $\Delta P$ on multiple models, suggesting that these approaches may degrade performance when applied to unsafe dialogue scenarios.

This gap highlights a fundamental limitation of existing prompt optimisation methods: they rely on static or search-based strategies that do not adapt to interaction context or optimisation history. In contrast, SafeCtrl-RL leverages a state-aware policy to dynamically select refinement strategies, enabling more effective and consistent behavioural control at inference time.

    \vspace{-0.2cm}
\subsection{Performance--Efficiency Ratio Analysis}

\begin{figure*}[ht]
\centering  
\includegraphics[width=.55\linewidth]{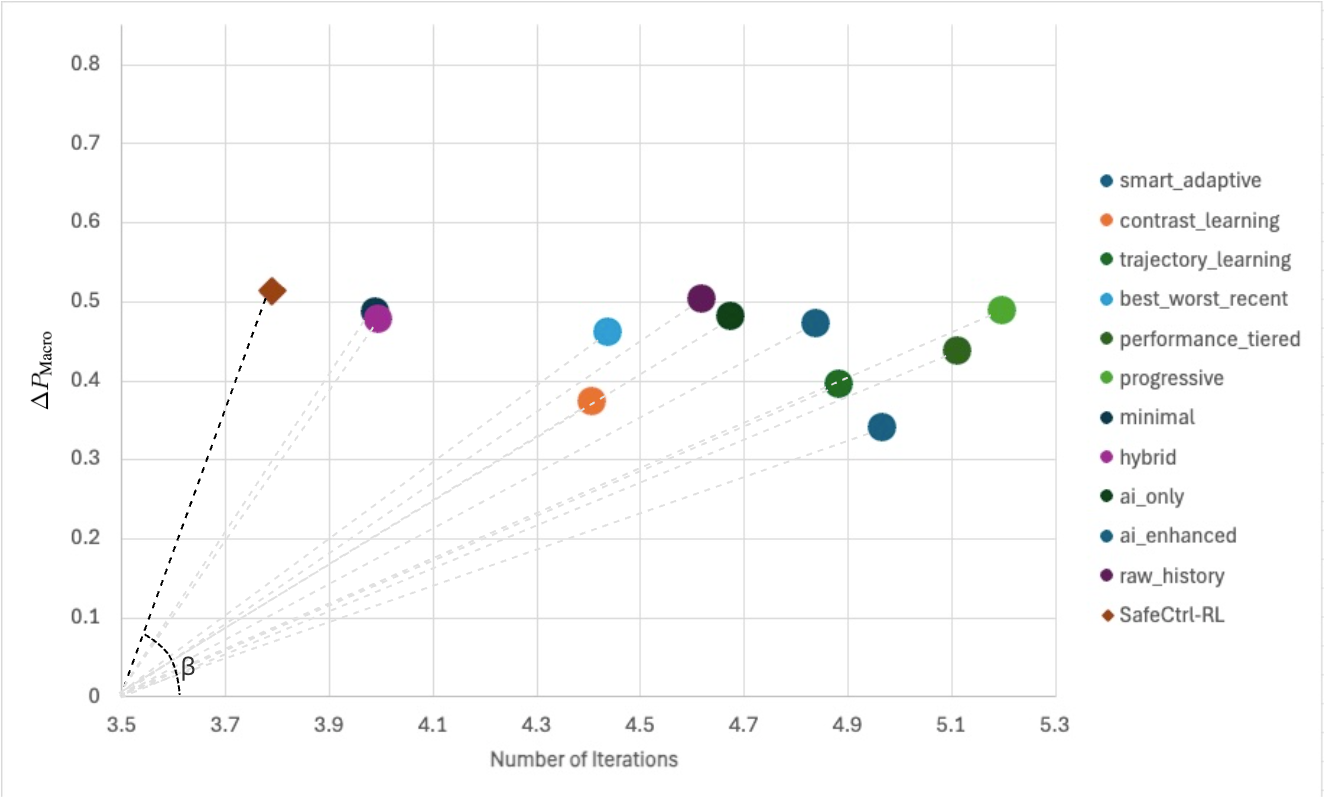}
\caption{Performance efficiency ratio, defined as the improvement in $\text{Macro-}P_{\text{Score}}$ from the baseline to the tested model ($\Delta P_{\text{Macro}}$) relative to the average number of iterations ($n$), i.e., $R_{\text{perf}} = \tan(\beta)$, for various hand-crafted strategies and SafeCtrl-RL.
\label{fig:performance_ratio}}
\end{figure*}

We analyse the trade-off between behavioural improvement ($\Delta P$) and computational cost (measured by the number of refinement iterations, $n$). Figure~\ref{fig:performance_ratio} plots $\Delta P$ against $n$ for all strategies, where steeper slopes correspond to higher Performance--Efficiency Ratios ($R_{\text{perf}}$), defined as the slope of the line connecting each point to the origin (0,0), where $\Delta P = 0$ and $n = 0$, such that $R_{\text{perf}} = \frac{\Delta P}{n} = \tan(\beta)$.

SafeCtrl-RL (brown marker) clearly occupies a favourable region of the space, achieving high improvement ($\Delta P \approx 0.51$) with relatively fewer iterations ($n \approx 3.7$). This results in a significantly steeper improvement trajectory compared to all handcrafted strategies, indicating superior efficiency. In contrast, most fixed strategies require more iterations ($n \approx 4.5$–$5.2$) to achieve comparable or lower improvements, leading to flatter slopes and reduced efficiency.

Notably, while several strategies such as \textit{progressive}, \textit{ai\_only}, and \textit{performance\_tiered} achieve moderately high $\Delta P$, they do so at substantially higher iteration counts, reflecting inefficient refinement processes. Other methods, such as \textit{contrast\_learning} and \textit{trajectory\_learning}, exhibit both lower improvement and higher cost, further reducing their overall effectiveness.

The dashed rays in the figure illustrate that most handcrafted strategies cluster along lower-efficiency directions, whereas SafeCtrl-RL forms a clear upper envelope. This suggests that adaptive strategy selection not only improves final performance but also accelerates convergence by selecting more effective refinement actions earlier in the optimisation process.

These results demonstrate that SafeCtrl-RL achieves a more favourable balance between improvement and computational cost, highlighting the advantage of policy-driven refinement over static heuristic strategies in efficiency-critical settings. \textbf{This indicates that SafeCtrl-RL achieves higher improvement per refinement step, effectively increasing optimisation efficiency by reducing unnecessary iterations.}

    \vspace{-0.15cm}
\subsection{Retention Analysis}

We evaluate whether behavioural improvements persist after removing SafeCtrl-RL, assessing the stability of induced behaviour beyond the refinement process. Table~\ref{tab:improvement_consistency} reports performance before safeguarding, after safeguarding, and after removing the control mechanism.

Across all models, SafeCtrl-RL yields substantial improvements over the plain baseline (e.g., Blacksheep: $0.193 \rightarrow 0.910$, DeepSeek-R1-Distill: $0.290 \rightarrow 0.793$). After removing the safeguard, performance decreases but remains consistently above the baseline, indicating that a significant portion of the improvement is retained.

The degree of retention varies across models. Evil-Alpaca achieves the highest retention (94.50\%), maintaining near-peak performance ($0.902$) after safeguard removal, suggesting strong intrinsic alignment and stability. Blacksheep also exhibits high retention (90.50\%) with a substantial residual gain ($+0.483$), indicating that improvements generalise beyond the control loop. DeepSeek-R1-Distill shows moderate retention (72.10\%) despite a large improvement ($+0.381$), implying partial persistence of refined behaviour. In contrast, DialoGPT demonstrates the weakest retention (60.12\%), with lower post-safeguard performance ($0.524$), suggesting greater reliance on external control.

These results indicate that inference-time control can induce partially sustained behavioural changes, effectively shaping response patterns beyond immediate optimisation. However, the extent of retention is model-dependent, highlighting the role of intrinsic alignment and steerability in sustaining improved behaviour.

\begin{table}[t]
\centering
\resizebox{\linewidth}{!}{%
\begin{tabular}{|l|c|c|c|c|c|}
\hline
\textbf{Model} & 
  \textbf{\begin{tabular}[c]{@{}c@{}} Plain Model \\ $P_{\text{Score}}^{\text{plain}}$ \end{tabular}} & 
  \textbf{\begin{tabular}[c]{@{}c@{}}SafeCtrl-RL \\ $P_{\text{Score}}^{\text{safeguarded}}$ \end{tabular}} & 
  \textbf{\begin{tabular}[c]{@{}c@{}}Post-Safeguard \\ $P_{\text{Score}}^{\text{post-safeguard}}$\end{tabular}} & 
  \textbf{\begin{tabular}[c]{@{}c@{}} $\Delta_{\text{mean}}$  \\ \end{tabular}} & 
  \textbf{\begin{tabular}[c]{@{}c@{}}Retention \\ \end{tabular}} \\ \hline \hline

\textbf{Blacksheep}        & 0.193 & 0.910 & 0.676 & +0.483 & 90.50\% \\ \hline
\textbf{Evil-Alpaca}       & 0.451 & 0.990 & 0.902 & +0.451 & 94.50\% \\ \hline
\textbf{DeepSeek-R1-Distill} & 0.290 & 0.793 & 0.671 & +0.381 & 72.10\% \\ \hline
\textbf{DialoGPT}          & 0.287 & 0.573 & 0.524 & +0.237 & 60.12\% \\ \hline
\end{tabular}%
}
\caption{Comparative performance ($P_{\text{Score}}$) and behavioural consistency metrics ($\Delta_{\text{mean}}$ is the pre–post safeguarding difference; \textbf{Retention} is the percentage of samples with reward $\geq 0.8$). (Further details in Appendix \ref{app:unlearning_experiment})}
\vspace{-0.4cm}
\label{tab:improvement_consistency}
\end{table}





\subsection{Discussion}

Across all experiments, SafeCtrl-RL demonstrates consistent advantages over both handcrafted and prompt optimisation methods in terms of effectiveness, robustness, and efficiency. The results show that no single prompt refinement strategy is universally optimal; instead, performance depends strongly on model characteristics and interaction context. By framing prompt selection as a sequential decision problem, SafeCtrl-RL dynamically adapts to these variations, enabling more stable and generalisable behaviour.

The efficiency analysis further highlights that adaptive control not only improves final performance but also accelerates convergence, achieving higher improvement per iteration compared to static strategies. In addition, the retention study suggests that inference-time control can induce partially sustained behavioural changes, although this effect varies across models.

These findings indicate that adaptive, policy-driven refinement is a more effective paradigm for inference-time safety control than fixed or search-based approaches. However, the framework relies on an external evaluator, which may introduce bias and latency constraints~\footnote{In the current implementation, latency is primarily dominated by external evaluation (see Appendix~\ref{app:latency}). Preliminary results with lightweight local evaluators reduce per-iteration latency to sub-second levels, suggesting that this bottleneck can be mitigated.}. This highlights the importance of developing more efficient and reliable feedback mechanisms for real-time deployment.

\section{Conclusion \& Future Work}

We introduced \textbf{SafeCtrl-RL}, an inference-time behavioural control framework that enables adaptive safety regulation in LLMs without parameter updates or retraining. By modelling prompt refinement as a sequential decision process, SafeCtrl-RL dynamically selects strategies based on context and feedback, achieving consistent improvements in both safety and response quality across diverse models. In addition to strong final performance, the framework demonstrates robust behavioural improvement and favourable performance--efficiency trade-offs, indicating effective optimisation with reduced refinement cost. The retention analysis further suggests that inference-time control can induce partially sustained behavioural changes.

Ongoing work focuses on improving efficiency and scalability. In particular, we aim to integrate \emph{offline} and \emph{online} adaptation by combining soft-prompting with SafeCtrl-RL, enabling the model to internalise common refinement patterns while retaining dynamic control. This hybrid approach has the potential to significantly reduce latency and iteration cost. In the future, we will extend the framework to multi-turn dialogue, multimodal LLMs, and domain-specific safety settings, as well as explore lightweight evaluators and alternative feedback signals to support real-time deployment.

\section{Limitation}

Although SafeCtrl-RL performs strongly across multiple settings, several limitations remain. First, our experiments focus primarily on \emph{under-aligned or unsafe} model families. While this setting highlights behavioural improvements, it remains unclear how the framework interacts with highly aligned, instruction-tuned foundation models. Although the method is model-agnostic, this requires further empirical validation.

Second, we evaluate only four LLM families. While these models vary in scale, alignment, and decoding behaviour, a broader evaluation across additional architectures would strengthen claims of generality and robustness.

Third, the behavioural retention analysis is based on a \emph{sample-driven evaluation}, using a limited subset of prompts across twelve safety categories. This provides an initial estimate of persistence but does not fully capture distribution-level effects. Future work will extend this analysis to the complete dataset to enable a more comprehensive assessment of long-term behavioural stability.

Finally, this study uses a single external evaluator, so the findings demonstrate the benefit of adaptive, policy-driven refinement within a fixed feedback regime, not evaluator-agnostic robustness. We keep the evaluator constant to isolate the effect of inference-time control and enable a controlled comparison. In this setting, the result is clear: adaptive refinement consistently outperforms fixed and search-based alternatives. Whether this holds for other evaluators is an important question for future work, but it does not affect the core comparative results reported here.





\bibliography{latex/new_references}

@inproceedings{pawelczyk2024incontextunlearninglanguagemodels,
author = {Pawelczyk, Martin and Neel, Seth and Lakkaraju, Himabindu},
title = {In-context unlearning: language models as few-shot unlearners},
year = {2024},
publisher = {JMLR.org},
booktitle = {Proceedings of the 41st International Conference on Machine Learning},
articleno = {1622},
numpages = {17},
location = {Vienna, Austria},
series = {ICML'24}
}

@inproceedings{
yang2024rlcdreinforcementlearningcontrastive,
title={{RLCD}: Reinforcement Learning from Contrastive Distillation for {LM} Alignment},
author={Kevin Yang and Dan Klein and Asli Celikyilmaz and Nanyun Peng and Yuandong Tian},
booktitle={The Twelfth International Conference on Learning Representations},
year={2024},
url={https://openreview.net/forum?id=v3XXtxWKi6}
}

@misc{zhang2024who,
      title={Who's Harry Potter? Approximate Unlearning in LLMs}, 
      author={Ronen Eldan and Mark Russinovich},
      year={2023},
      eprint={2310.02238},
      archivePrefix={arXiv},
      primaryClass={cs.CL},
      url={https://arxiv.org/abs/2310.02238}, 
}

@inproceedings{bender2021dangers,
author = {Bender, Emily M. and Gebru, Timnit and McMillan-Major, Angelina and Shmitchell, Shmargaret},
title = {On the Dangers of Stochastic Parrots: Can Language Models Be Too Big? },
year = {2021},
isbn = {9781450383097},
publisher = {Association for Computing Machinery},
address = {New York, NY, USA},
url = {https://doi.org/10.1145/3442188.3445922},
doi = {10.1145/3442188.3445922},
booktitle = {Proceedings of the 2021 ACM Conference on Fairness, Accountability, and Transparency},
pages = {610–623},
numpages = {14},
location = {Virtual Event, Canada},
series = {FAccT '21}
}

@article{yao2023editing,
author = {Wang, Song and Zhu, Yaochen and Liu, Haochen and Zheng, Zaiyi and Chen, Chen and Li, Jundong},
title = {Knowledge Editing for Large Language Models: A Survey},
year = {2024},
issue_date = {March 2025},
publisher = {Association for Computing Machinery},
address = {New York, NY, USA},
volume = {57},
number = {3},
issn = {0360-0300},
url = {https://doi.org/10.1145/3698590},
doi = {10.1145/3698590},
journal = {ACM Comput. Surv.},
month = nov,
articleno = {59},
numpages = {37}
}

@article{brown2020language,
  title={Language models are few-shot learners},
  author={Brown, Tom and Mann, Benjamin and Ryder, Nick and Subbiah, Melanie and Kaplan, Jared D and Dhariwal, Prafulla and Neelakantan, Arvind and Shyam, Pranav and Sastry, Girish and Askell, Amanda and others},
  journal={Advances in neural information processing systems},
  volume={33},
  pages={1877--1901},
  year={2020}
}

@inproceedings{zhang2020dialogpt,
  title={Dialogpt: Large-scale generative pre-training for conversational response generation},
  author={Zhang, Yizhe and Sun, Siqi and Galley, Michel and Chen, Yen-Chun and Brockett, Chris and Gao, Xiang and Gao, Jianfeng and Liu, Jingjing and Dolan, William B},
  booktitle={Proceedings of the 58th annual meeting of the association for computational linguistics: system demonstrations},
  pages={270--278},
  year={2020}
}

@misc{deepseekai2025deepseek,
      title={DeepSeek-R1: Incentivizing Reasoning Capability in LLMs via Reinforcement Learning}, 
      author={DeepSeek-AI and collaborators},
      year={2025},
      eprint={2501.12948},
      archivePrefix={arXiv},
      primaryClass={cs.CL},
      url={https://arxiv.org/abs/2501.12948}, 
}

@article{unlearning_survey,
author = {Nguyen, Thanh Tam and Huynh, Thanh Trung and Ren, Zhao and Nguyen, Phi Le and Liew, Alan Wee-Chung and Yin, Hongzhi and Nguyen, Quoc Viet Hung},
title = {A Survey of Machine Unlearning},
year = {2025},
issue_date = {October 2025},
publisher = {Association for Computing Machinery},
address = {New York, NY, USA},
volume = {16},
number = {5},
issn = {2157-6904},
url = {https://doi.org/10.1145/3749987},
doi = {10.1145/3749987},
journal = {ACM Trans. Intell. Syst. Technol.},
month = sep,
articleno = {108},
numpages = {46}
}

@inproceedings{ethayarajh2024kto,
author = {Ethayarajh, Kawin and Xu, Winnie and Muennighoff, Niklas and Jurafsky, Dan and Kiela, Douwe},
title = {Model alignment as prospect theoretic optimization},
year = {2024},
publisher = {JMLR.org},
booktitle = {Proceedings of the 41st International Conference on Machine Learning},
articleno = {504},
numpages = {18},
location = {Vienna, Austria},
series = {ICML'24}
}

@inproceedings{burns2023weaktostrong,
author = {Burns, Collin and Izmailov, Pavel and Kirchner, Jan Hendrik and Baker, Bowen and Gao, Leo and Aschenbrenner, Leopold and Chen, Yining and Ecoffet, Adrien and Joglekar, Manas and Leike, Jan and Sutskever, Ilya and Wu, Jeff},
title = {Weak-to-strong generalization: eliciting strong capabilities with weak supervision},
year = {2024},
publisher = {JMLR.org},
booktitle = {Proceedings of the 41st International Conference on Machine Learning},
articleno = {196},
numpages = {42},
location = {Vienna, Austria},
series = {ICML'24}
}

@techreport{LLMsInEducation,
title = "The Revolution Has Arrived: What the Current State of Large Language Models in Education Implies for the Future",
author = "Russell Beale",
year = "2025",
month = jul,
day = "2",
doi = "10.48550/arXiv.2507.02180",
language = "English",
publisher = "arXiv",
type = "WorkingPaper",
institution = "arXiv",
}

@misc{cui2024risk,
      title={Risk Taxonomy, Mitigation, and Assessment Benchmarks of Large Language Model Systems}, 
      author={Tianyu Cui and Yanling Wang and Chuanpu Fu and Yong Xiao and Sijia Li and Xinhao Deng and Yunpeng Liu and Qinglin Zhang and Ziyi Qiu and Peiyang Li and Zhixing Tan and Junwu Xiong and Xinyu Kong and Zujie Wen and Ke Xu and Qi Li},
      year={2024},
      eprint={2401.05778},
      archivePrefix={arXiv},
      primaryClass={cs.CL},
      url={https://arxiv.org/abs/2401.05778}, 
}

@inproceedings{meng2022locating,
  title = {Locating and Editing Factual Associations in GPT},
  author = {Kevin Meng and David Bau and Alex Andonian and Yonatan Belinkov},
  booktitle = {NeurIPS 2022},
  year = {2022},
  location = {New Orleans, LA, USA},

}

@inproceedings{meng2023massediting,
  title = {Mass-Editing Memory in a Transformer},
  author = {Kevin Meng and Samuel Mendelsohn and David Bau and Yonatan Belinkov},
  booktitle = {International Conference on Learning Representations (ICLR 2023)},
  year = {2023},
  location = {Kigali, Rwanda}
}

@inproceedings{wang2023voyager,
  title = {Voyager: An Open-Ended Embodied Agent with Large Language Models},
  author = {Guanzhi Wang and Yuqi Xie and Yunfan Jiang and Ajay Mandlekar and Chaowei Xiao and Yuke Zhu and Linxi Fan and Anima Anandkumar},
  booktitle = {Proceedings of the 41st International Conference on Machine Learning (ICML 2024)},
  series = {Proceedings of Machine Learning Research},
  volume = {235},
  year = {2024},
  location = {Vienna, Austria}
}

@article{bran2023chemcro,
  author    = {Andres M. Bran and Sam Cox and Oliver Schilter and Carlo Baldassari and Andrew D. White and Philippe Schwaller},
  title     = {Augmenting large language models with chemistry tools},
  journal   = {Nature Machine Intelligence},
  year      = {2024},
  volume    = {6},
  number    = {5},
  pages     = {525--535},
  doi       = {10.1038/s42256-024-00832-8},
  url       = {https://doi.org/10.1038/s42256-024-00832-8},
  issn      = {2522-5839}
}

@inproceedings{gupta2024unifiedframeworkmodelediting,
  title={A unified framework for model editing},
  author={Gupta, Akshat and Sajnani, Dev and Anumanchipalli, Gopala},
  booktitle={Findings of the Association for Computational Linguistics: EMNLP 2024},
  pages={15403--15418},
  year={2024}
}

@inproceedings{
jiang2025anyedit,
title={AnyEdit: Edit Any Knowledge Encoded in Language Models},
author={Houcheng Jiang and Junfeng Fang and Ningyu Zhang and Mingyang Wan and Guojun Ma and Xiang Wang and Xiangnan He and Tat-Seng Chua},
booktitle={Forty-second International Conference on Machine Learning},
year={2025},
url={https://openreview.net/forum?id=aJIoBur0Ef}
}

@misc{liu2025promptcontentenhancingllm,
      title={Beyond Prompt Content: Enhancing LLM Performance via Content-Format Integrated Prompt Optimization}, 
      author={Yuanye Liu and Jiahang Xu and Li Lyna Zhang and Qi Chen and Xuan Feng and Yang Chen and Zhongxin Guo and Yuqing Yang and Peng Cheng},
      year={2025},
      eprint={2502.04295},
      archivePrefix={arXiv},
      primaryClass={cs.CL},
      url={https://arxiv.org/abs/2502.04295}, 
}

@inproceedings{das2025greater,
title={{GR}eaTer: Gradients Over Reasoning Makes Smaller Language Models Strong Prompt Optimizers},
author={Sarkar Snigdha Sarathi Das and Ryo Kamoi and Bo Pang and Yusen Zhang and Caiming Xiong and Rui Zhang},
booktitle={The Thirteenth International Conference on Learning Representations},
year={2025},
url={https://openreview.net/forum?id=fWRBheSJth}
}

@inproceedings{ji2025pku,
  title={Pku-saferlhf: Towards multi-level safety alignment for llms with human preference},
  author={Ji, Jiaming and Hong, Donghai and Zhang, Borong and Chen, Boyuan and Dai, Josef and Zheng, Boren and Qiu, Tianyi Alex and Zhou, Jiayi and Wang, Kaile and Li, Boxun and others},
  booktitle={Proceedings of the 63rd Annual Meeting of the Association for Computational Linguistics (Volume 1: Long Papers)},
  pages={31983--32016},
  year={2025}
}

@article{ji2023beavertails,
  title={Beavertails: Towards improved safety alignment of llm via a human-preference dataset},
  author={Ji, Jiaming and Liu, Mickel and Dai, Josef and Pan, Xuehai and Zhang, Chi and Bian, Ce and Chen, Boyuan and Sun, Ruiyang and Wang, Yizhou and Yang, Yaodong},
  journal={Advances in Neural Information Processing Systems},
  volume={36},
  pages={24678--24704},
  year={2023}
}

@article{rafailov2023direct,
  title={Direct preference optimization: Your language model is secretly a reward model},
  author={Rafailov, Rafael and Sharma, Archit and Mitchell, Eric and Manning, Christopher D and Ermon, Stefano and Finn, Chelsea},
  journal={Advances in neural information processing systems},
  volume={36},
  pages={53728--53741},
  year={2023}
}

@misc{yang2024largelanguagemodelsoptimizers,
      title={Large Language Models as Optimizers}, 
      author={Chengrun Yang and Xuezhi Wang and Yifeng Lu and Hanxiao Liu and Quoc V. Le and Denny Zhou and Xinyun Chen},
      year={2024},
      eprint={2309.03409},
      archivePrefix={arXiv},
      primaryClass={cs.LG},
      url={https://arxiv.org/abs/2309.03409}, 
}

@inproceedings{prasad2023gripsgradientfreeeditbasedinstruction,
  title={Grips: Gradient-free, edit-based instruction search for prompting large language models},
  author={Prasad, Archiki and Hase, Peter and Zhou, Xiang and Bansal, Mohit},
  booktitle={Proceedings of the 17th Conference of the European Chapter of the Association for Computational Linguistics},
  pages={3845--3864},
  year={2023}
}

@inproceedings{tang2025unleashingpotentiallargelanguage,
  title={Unleashing the potential of large language models as prompt optimizers: Analogical analysis with gradient-based model optimizers},
  author={Tang, Xinyu and Wang, Xiaolei and Zhao, Wayne Xin and Lu, Siyuan and Li, Yaliang and Wen, Ji-Rong},
  booktitle={Proceedings of the AAAI Conference on Artificial Intelligence},
  volume={39},
  number={24},
  pages={25264--25272},
  year={2025}
}

@article{wei2022chain,
  title={Chain-of-thought prompting elicits reasoning in large language models},
  author={Wei, Jason and Tay, Yi and Bommasani, Rishi and Ritter, Michael and Macherey, Derek and Chung, Quoc V and Houlsby, Neil and Luong, Thang},
  journal={Advances in Neural Information Processing Systems},
  volume={35},
  pages={24824--24837},
  year={2022}
}

@inproceedings{madaan2023selfcorrect,
author = {Madaan, Aman and Tandon, Niket and Gupta, Prakhar and Hallinan, Skyler and Gao, Luyu and Wiegreffe, Sarah and Alon, Uri and Dziri, Nouha and Prabhumoye, Shrimai and Yang, Yiming and Gupta, Shashank and Majumder, Bodhisattwa Prasad and Hermann, Katherine and Welleck, Sean and Yazdanbakhsh, Amir and Clark, Peter},
title = {SELF-REFINE: iterative refinement with self-feedback},
year = {2023},
publisher = {Curran Associates Inc.},
address = {Red Hook, NY, USA},
booktitle = {Proceedings of the 37th International Conference on Neural Information Processing Systems},
articleno = {2019},
numpages = {61},
location = {New Orleans, LA, USA},
series = {NIPS '23}
}

@article{kojima2022large,
  title={Large language models are zero-shot reasoners},
  author={Kojima, Takeshi and Gu, Shixiang Shane and Reid, Miltos and Matsuo, Yutaka and Gu, Scott},
  journal={Advances in Neural Information Processing Systems},
  volume={35},
  pages={32247--32257},
  year={2022}
}

@article{ouyang2022training,
  title={Training language models to follow instructions with human feedback},
  author={Ouyang, Long and Wu, Jeff and Jiang, Xu and Almeida, Diogo and Wainwright, Carroll L and Mishkin, Pamela and Zhang, Chong and Agarwal, Sandeep and Slama, Kalian and Ray, Tony and others},
  journal={Advances in neural information processing systems},
  volume={35},
  pages={27730--27744},
  year={2022}
}

@article{touvron2023llama,
  title={Llama 2: Open foundation and chat models},
  author={Touvron, Hugo and Martin, Louis and Stone, Kevin and Albert, Peter and Almahairi, Amjad and Babaei, Yasmine and Bashlykov, Nikolay and Batra, Soumya and Bhargava, Prajjwal and Bhosale, Shruti and others},
  journal={arXiv preprint arXiv:2307.09288},
  year={2023}
}

@inproceedings{rubin2022learning,
  title={Learning to Retrieve Prompts for In-Context Learning},
  author={Rubin, Ohad and Herzig, Jonathan and Berant, Jonathan},
  booktitle={Proceedings of the 2022 Conference of the North American Chapter of the Association for Computational Linguistics: Human Language Technologies},
  pages={3755--3771},
  year={2022}
}

@article{liu2024textgrad,
  title={Textgrad: Automatic differentiation via text},
  author={Liu, Zhengzhong and Groth, Oliver and Maitland, Adam and Gasic, Milica and Ermon, Stefano and others},
  journal={arXiv preprint arXiv:2406.12614},
  year={2024}
}

@inproceedings{
guo2023connecting,
title={Connecting Large Language Models with Evolutionary Algorithms Yields Powerful Prompt Optimizers},
author={Qingyan Guo and Rui Wang and Junliang Guo and Bei Li and Kaitao Song and Xu Tan and Guoqing Liu and Jiang Bian and Yujiu Yang},
booktitle={The Twelfth International Conference on Learning Representations},
year={2024},
url={https://openreview.net/forum?id=ZG3RaNIsO8}
}

@inproceedings{deng-etal-2022-rlprompt,
    title = "{RLP}rompt: Optimizing Discrete Text Prompts with Reinforcement Learning",
    author = "Deng, Mingkai  and
      Wang, Jianyu  and
      Hsieh, Cheng-Ping  and
      Wang, Yihan  and
      Guo, Han  and
      Shu, Tianmin  and
      Song, Meng  and
      Xing, Eric  and
      Hu, Zhiting",
    editor = "Goldberg, Yoav  and
      Kozareva, Zornitsa  and
      Zhang, Yue",
    booktitle = "Proceedings of the 2022 Conference on Empirical Methods in Natural Language Processing",
    month = dec,
    year = "2022",
    address = "Abu Dhabi, United Arab Emirates",
    publisher = "Association for Computational Linguistics",
    url = "https://aclanthology.org/2022.emnlp-main.222/",
    doi = "10.18653/v1/2022.emnlp-main.222",
    pages = "3369--3391"
}

@inproceedings{zhou2022large,
  title={Large language models are human-level prompt engineers},
  author={Zhou, Yongchao and Muresanu, Andrei Ioan and Han, Ziwen and Paster, Keiran and Pitis, Silviu and Chan, Harris and Ba, Jimmy},
  booktitle={The eleventh international conference on learning representations},
  year={2022}
}

@inproceedings{tang2025unleashing,
  title={Unleashing the potential of large language models as prompt optimizers: Analogical analysis with gradient-based model optimizers},
  author={Tang, Xinyu and Wang, Xiaolei and Zhao, Wayne Xin and Lu, Siyuan and Li, Yaliang and Wen, Ji-Rong},
  booktitle={Proceedings of the AAAI Conference on Artificial Intelligence},
  volume={39},
  number={24},
  pages={25264--25272},
  year={2025}
}

@inproceedings{fernandopromptbreeder,
  title={Promptbreeder: Self-Referential Self-Improvement via Prompt Evolution},
  author={Fernando, Chrisantha and Banarse, Dylan Sunil and Michalewski, Henryk and Osindero, Simon and Rockt{\"a}schel, Tim},
  booktitle={Forty-first International Conference on Machine Learning},
  year={2023}
}

@article{choi2025efficient,
  title={Efficient Prompt Optimization for Relevance Evaluation via LLM-Based Confusion Matrix Feedback},
  author={Choi, Jaekeol},
  journal={Applied Sciences},
  volume={15},
  number={9},
  pages={5198},
  year={2025},
  publisher={MDPI}
}

@inproceedings{
khattab2024dspy,
title={{DSP}y: Compiling Declarative Language Model Calls into State-of-the-Art Pipelines},
author={Omar Khattab and Arnav Singhvi and Paridhi Maheshwari and Zhiyuan Zhang and Keshav Santhanam and Sri Vardhamanan A and Saiful Haq and Ashutosh Sharma and Thomas T. Joshi and Hanna Moazam and Heather Miller and Matei Zaharia and Christopher Potts},
booktitle={The Twelfth International Conference on Learning Representations},
year={2024},
url={https://openreview.net/forum?id=sY5N0zY5Od}
}

@inproceedings{stiennon2020learning,
  title={Learning to Summarize with Human Feedback},
  author={Stiennon, Nisan and Ouyang, Long and Wu, Jeff and others},
  booktitle={Advances in Neural Information Processing Systems (NeurIPS)},
  volume={33},
  pages={3008--3021},
  year={2020}
}

@misc{bai2022constitutionalaiharmlessnessai,
      title={Constitutional AI: Harmlessness from AI Feedback}, 
      author={Yuntao Bai and Saurav Kadavath and Sandipan Kundu and Amanda Askell and Jackson Kernion and Andy Jones and Anna Chen and Anna Goldie and Azalia Mirhoseini and Cameron McKinnon and Carol Chen and Catherine Olsson and Christopher Olah and Danny Hernandez and Dawn Drain and Deep Ganguli and Dustin Li and Eli Tran-Johnson and Ethan Perez and Jamie Kerr and Jared Mueller and Jeffrey Ladish and Joshua Landau and Kamal Ndousse and Kamile Lukosuite and Liane Lovitt and Michael Sellitto and Nelson Elhage and Nicholas Schiefer and Noemi Mercado and Nova DasSarma and Robert Lasenby and Robin Larson and Sam Ringer and Scott Johnston and Shauna Kravec and Sheer El Showk and Stanislav Fort and Tamera Lanham and Timothy Telleen-Lawton and Tom Conerly and Tom Henighan and Tristan Hume and Samuel R. Bowman and Zac Hatfield-Dodds and Ben Mann and Dario Amodei and Nicholas Joseph and Sam McCandlish and Tom Brown and Jared Kaplan},
      year={2022},
      eprint={2212.08073},
      archivePrefix={arXiv},
      primaryClass={cs.CL},
      url={https://arxiv.org/abs/2212.08073}, 
}

\appendix

\section{Appendix}
\subsection{LLM Usage Disclosure}
Grammarly was used to aid in the writing of this paper. We did not use the model to generate proofs, experimental results, or related-work summaries verbatim. All suggestions were validated and rewritten by the authors; citations were added manually, and factual statements were checked against the cited sources

\section{Full version of the Related Work}\label{sec:related_full}

Controlling the behaviour of large language models (LLMs) is an increasingly important problem in dialogue systems, where safety, coherence, and contextual appropriateness must be maintained across evolving interactions. Existing approaches can be broadly categorised into parameter-based editing and prompt-based control.

\paragraph{Parameter-Based Editing.}
Parameter-editing methods aim to modify internal model representations to update or remove specific knowledge. Approaches such as ROME~\citep{meng2022locating} and MEMIT~\citep{meng2023massediting} identify and alter localised “knowledge neurons,” while more recent frameworks~\citep{yao2023editing, jiang2025anyedit, gupta2024unifiedframeworkmodelediting} provide systematic techniques for large-scale model editing. These methods are effective in controlled settings and enable precise manipulation of model behaviour. However, they typically require white-box access to model parameters, operate offline, and may introduce unintended global side effects. As a result, they are not well suited to interactive dialogue scenarios, where behaviour must be adjusted dynamically during ongoing conversations.

\paragraph{Prompt-Based Unlearning and Behaviour Steering.} 
Prompt-based methods offer a more flexible alternative by steering behaviour through input conditioning rather than parameter modification. In-context unlearning embeds corrective instructions directly into prompts to suppress undesirable outputs~\citep{pawelczyk2024incontextunlearninglanguagemodels, zhang2024who}, building on the broader paradigm of in-context learning~\citep{brown2020language}. These approaches enable inference-time intervention and are compatible with black-box models, making them suitable for real-world deployment.

A wide range of static prompting strategies has been explored, including few-shot prompting, role conditioning, chain-of-thought reasoning~\citep{wei2022chain}, value reinforcement inspired by RLHF~\citep{ouyang2022training}, and constitutional prompting~\citep{bai2022constitutionalaiharmlessnessai}. These methods encode safety constraints through fixed prompt structures and typically operate as \emph{single-pass transformations}. While effective in some cases, their performance is often model-dependent and lacks robustness across scenarios, as also observed in our experiments.

An important class of methods introduces \emph{iterative refinement} through generate--revise paradigms, most notably \textit{Self-Correction}~\citep{madaan2023selfcorrect}, where model outputs are refined via a second-pass prompt. This mechanism enables multi-step behaviour adjustment and is more compatible with iterative interaction settings. However, these approaches remain static, relying on fixed revision strategies without adapting to feedback or optimisation history.

These approaches enable inference-time intervention and are compatible with black-box models. However, they typically rely on static prompts, limiting their ability to adapt to changing conversational context. This limitation is particularly pronounced in dialogue settings, where safety issues often arise from context accumulation across multiple turns rather than single inputs.

\paragraph{Prompt Optimisation and Iterative Refinement.}
To address this, recent work has explored prompt optimisation and iterative refinement strategies. Methods such as OPRO~\citep{yang2024largelanguagemodelsoptimizers}, GPO~\citep{tang2025unleashingpotentiallargelanguage}, and GRIPS~\citep{prasad2023gripsgradientfreeeditbasedinstruction} use external models or search-based techniques to improve prompts, while RL-inspired approaches incorporate optimisation signals for alignment~\citep{yang2024rlcdreinforcementlearningcontrastive, liu2025promptcontentenhancingllm, das2025greater}. Although these methods introduce iterative processes, they typically produce optimised prompts offline or apply fixed refinement strategies at deployment time. Consequently, they lack the ability to adapt their behaviour dynamically based on evolving dialogue context and interaction history.

\begin{table*}[t]
\centering
\small
\begin{tabular}{l|c|c|c}
\hline
\textbf{Approach} & \textbf{Method Type} & \textbf{Refinement Type} & \textbf{Dialogue-Aware} \\
\hline
ROME~\citep{meng2022locating} & Parameter-based & One-shot & $\times$ \\
MEMIT~\citep{meng2023massediting} & Parameter-based & One-shot & $\times$ \\
Model Editing~\citep{yao2023editing, jiang2025anyedit} & Parameter-based & One-shot & $\times$ \\
Few-Shot Prompting~\citep{brown2020language} & Prompt-based & One-shot & $\times$ \\
Role Conditioning~\citep{touvron2023llama} & Prompt-based & One-shot & $\times$ \\
Chain-of-Thought~\citep{wei2022chain} & Prompt-based & One-shot & $\times$ \\
Value Reinforcement~\citep{ouyang2022training} & Prompt-based & One-shot & $\times$ \\
Constitutional Prompting~\citep{bai2022constitutionalaiharmlessnessai} & Prompt-based & One-shot & $\times$ \\
Basic Prompt Injection & Prompt-based & One-shot & $\times$ \\
Self-Correction~\citep{madaan2023selfcorrect} & Prompt-based & Iterative & $\times$ \\
Enhanced Self-Correction~\citep{bai2022constitutionalaiharmlessnessai} & Prompt-based & Iterative & $\times$ \\
OPRO~\citep{yang2024largelanguagemodelsoptimizers} & Prompt-based & Iterative (offline) & $\times$ \\
GRIPS~\citep{prasad2023gripsgradientfreeeditbasedinstruction} & Prompt-based & Iterative (offline) & $\times$ \\
TextGrad~\citep{liu2024textgrad} & Prompt-based & Iterative (feedback) & Partial \\
Evolutionary Prompting~\citep{guo2023connecting} & Prompt-based & Iterative (search) & $\times$ \\
Dynamic Retrieval~\citep{rubin2022learning} & Prompt-based & Iterative (heuristic) & Partial \\
\hline
\end{tabular}
\caption{Comparison of behaviour control approaches. Methods are categorised by type, refinement strategy, and whether they adapt to dialogue context.}
\label{tab:taxonomy_comparison}
\end{table*}

Table~\ref{tab:taxonomy_comparison} summarises the key properties of existing behaviour control approaches across method type, refinement strategy, and dialogue awareness. Parameter-based methods provide precise control but require model access and operate offline, limiting their applicability in real-world settings. Prompt-based approaches are more deployable, but most rely on either one-shot prompting or fixed iterative strategies, without adapting to interaction context. While prompt optimisation methods introduce iterative refinement, they typically operate offline or apply static optimisation strategies at deployment, lacking explicit modelling of sequential decision-making during dialogue.

In contrast, we introduce \textbf{SafeCtrl-RL}, which formulates safety as an inference-time control problem and enables adaptive prompt refinement based on safety--quality feedback and conversational context. By integrating reinforcement learning with a structured representation of interaction history, SafeCtrl-RL dynamically selects refinement strategies during interaction, allowing behaviour to be continuously adjusted in response to evolving dialogue without retraining or parameter access.

\section{Detailed SafeCtrl-RL Setup}

\subsection{Experimental Environment}
\label{sec:exp_env}
The experiments were carried out on Google Cloud Compute Engine using a virtual machine equipped with 12 Intel Cascade Lake vCPUs, 48 GB of RAM, and an NVIDIA L4 GPU. Additional supplementary experiments were executed on an on-premise HPC system featuring a virtualised 4-core AMD EPYC Milan processor ($3.0$ GHz) with $15$ GB of DDR4 memory. The HPC environment also included four NVIDIA A100 PCIe GPUs, each with $80$ GB of HBM2e memory, to support large-scale model evaluation. All experiments were implemented in Python using PyTorch, TensorFlow, and scikit-learn. The HPC server ran Red Hat Enterprise Linux 8.10 with kernel version 4.18.


\subsection{State Vector}
\label{app:state_vector}

\begin{table*}[th!]
\centering
\resizebox{0.9\textwidth}{!}{%
\begin{tabularx}{\textwidth}{@{}clX@{}}
\toprule
\textbf{Index} & \textbf{Feature} & \textbf{Description} \\ 
\midrule
\multicolumn{3}{l}{\textbf{1. Training Progress Features (1--5)}} \\
1 & Category Progress & Progress through categories, normalised $[0,1]$. \\
2 & Prompt-in-Category Progress & Progress within current category prompts $[0,1]$. \\
3 & Episode Progress & Episodes completed, normalised to 200 $[0,1]$. \\
4 & Performance History Fullness & Fraction of history buffer used (50 max). \\
5 & Cache Utilisation & Fraction of cache filled (2,000 max). \\
\midrule

\multicolumn{3}{l}{\textbf{2. Performance Features (6--11)}} \\
6 & Recent Mean Performance & Mean of the most recent up to 10 scores. \\
7 & Recent Volatility & Standard deviation of recent scores. \\
8 & Overall Mean Performance & Mean across all recorded scores. \\
9 & Performance Trend & Linear slope over the most recent 5 scores. \\
10 & Improvement Ratio & Recent mean vs. early mean (first 5 scores). \\
11 & Performance Consistency & $1 / (1 + \text{std})$ \\
\midrule

\multicolumn{3}{l}{\textbf{3. Strategy Performance Features (12--15)}} \\
12 & Best Strategy Performance & Maximum mean score among strategies. \\
13 & Strategy Diversity & Std.\ of strategy performance scores. \\
14 & Strategy Usage Balance & Evenness of strategy usage distribution. \\
15 & Total Strategy Experience & Normalised count of strategy uses (100 max). \\
\midrule

\multicolumn{3}{l}{\textbf{4. Current Episode State Features (16--19)}} \\
16 & Iteration Progress & Iterations done / max per episode. \\
17 & Best Reward This Episode & Best reward in episode so far. \\
18 & Episode Difficulty & Estimated complexity of current prompt. \\
19 & Episode Momentum & The difference in rewards between the first and last attempts. \\
\midrule

\multicolumn{3}{l}{\textbf{5. Category and Context Features (20--24)}} \\
20 & Category Feature 0 & Normalised number of prompts in category. \\
21 & Category Feature 1 & Average prompt length (chars), normalised. \\
22 & Category Feature 2 & Std.\ of prompt length, normalised. \\
23 & Category Feature 3 & Average word count per prompt, normalised. \\
24 & Category Feature 4 & Complexity score (questions, “why”, “explain”), normalised. \\
\midrule

\multicolumn{3}{l}{\textbf{6. Risk and Safety Features (25--27)}} \\
25 & Unsafe Category Score & Encoded safety risk of category. \\
26 & Recent Error Rate & Fraction of recent 10 experiences that were errors. \\
27 & Consecutive Errors & Normalised streak of consecutive errors. \\
\midrule

\multicolumn{3}{l}{\textbf{7. Exploration State Features (28--30)}} \\
28 & Exploration Rate & Current $\epsilon$ (exploration factor). \\
29 & Scaled Learning Rate & Learning rate normalised. \\
30 & Convergence Indicator & $1$ if recent std. $< 0.05$, else $0$. \\
\midrule

\multicolumn{3}{l}{\textbf{8. Advanced Prompt Features (31--32)}} \\
31 & Prompt Sophistication & Fraction of sophistication keywords present. \\
32 & Instruction Density & Instruction-related words per prompt length. \\
\midrule

\multicolumn{3}{l}{\textbf{9. Hash-based User Prompt Features (33--36)}} \\
33 & User Prompt Hash 0 & Hash fingerprint feature 0. \\
34 & User Prompt Hash 1 & Hash fingerprint feature 1. \\
35 & User Prompt Hash 2 & Hash fingerprint feature 2. \\
36 & User Prompt Hash 3 & Hash fingerprint feature 3. \\
\bottomrule
\end{tabularx}
}
\caption{Structure of the SafeCtrl-RL state vector (36 total)}
\label{tab:obs}
\end{table*}

To enable adaptive and context-aware strategy selection, the SafeCtrl-RL agent encodes each refinement step as a fixed-length state vector. This state representation aggregates information from the current interaction, historical optimisation trajectories, and internal learning dynamics of the agent. All features are normalised to ensure numerical stability and comparability across episodes, and are derived deterministically from observable signals during execution.

Table~\ref{tab:obs} details the full structure of the 36-dimensional state vector used by the RL agent. Features are organised into nine semantically coherent groups, capturing Training progress, Recent and long-term performance statistics, Strategy effectiveness, Episode-level dynamics, Category-specific context, Safety risk signals, Exploration state, Prompt-level complexity, and a Lightweight fingerprint of the user input. Together, these features provide a compact yet expressive summary of the safeguard’s operational context, enabling the agent to reason over both immediate safety outcomes and longer-term behavioural trends.

\subsubsection{Example}
\label{sec:state_example}

To illustrate the deterministic mapping from raw environment logs to the 36-dimensional input vector, Table~\ref{tab:worked_example} provides a snapshot of the agent's state during an active training session. In this example, the agent is navigating a high-complexity prompt in category \texttt{S10}.

\begin{table*}[ht]
\centering
\begin{small}
\begin{tabularx}{\textwidth}{l c l X}
\toprule
\textbf{Feature Group} & \textbf{Index} & \textbf{Value} & \textbf{Operational Context} \\ \midrule
\textbf{Progress}    & 1      & 0.833  & 10/12 categories completed. \\
\textbf{Performance} & 6      & 0.107  & Low mean reward (stagnation signal). \\
\textbf{Volatility}  & 7      & 0.099  & Stable performance variance. \\
\textbf{Trend}       & 9      & -0.042 & Negative slope; signals current strategy is failing. \\
\textbf{Consistency} & 11     & 0.910  & High score stability (1 is good consistency). \\
\textbf{Exp. Factor} & 15     & 0.700  & Significant strategy experience . \\
\textbf{Complexity}  & 18     & 0.540  & Moderate prompt difficulty estimate. \\
\textbf{Risk}        & 25     & 0.500  & \texttt{S10} category assigned a medium risk weight. \\
\textbf{Safety}      & 26     & 0.000  & Perfect safety record in last 10 attempts. \\
\textbf{Exploration} & 28     & 0.200  & Agent is in exploitation-heavy phase ($\epsilon=0.2$). \\
\textbf{Semantic}    & 31     & 0.143  & Low keyword sophistication (1/7 matches). \\
\textbf{Density}     & 32     & 0.149  & High word count relative to instructions. \\
\textbf{Hash}        & 33--36 & [0.12, \dots] & Latent vector for prompt similarity recognition. \\ \bottomrule
\end{tabularx}
\end{small}
\caption{Worked Example of History-to-State Mapping (Snapshot at Episode 11). Indices 2-5, 10, 12-14, 16, 17, 19, 20-24, 27, 29, 30 all have value zero.}
\label{tab:worked_example}
\end{table*}

\subsection{DQN Parameters}
\label{app:DQN-params}

Table~\ref{tab:DQN-Params} details the hyperparameter configuration of the Deep Q-Network (DQN) employed in both training and experimental evaluation.

\begin{itemize}
    \item \textbf{Network Architecture $[512, 512, 256]$}: A high-capacity Multi-Layer Perceptron (MLP). The depth allows the model to approximate complex, non-linear mappings between the state space and action-value space .
    \item \textbf{Learning Rate ($\alpha = 1.0 \times 10^{-4}$)}: A small learning rate is chosen to ensure smooth updates to the weights $\theta$.
    \item \textbf{Batch Size ($N=256$)}: By sampling $256$ transitions from the replay buffer $\mathcal{D}$, we reduce the variance of the stochastic gradient descent updates, leading to more stable optimisation.
    \item \textbf{Exploration Fraction}: We employ an $\epsilon$-greedy strategy, where $\epsilon$ decays linearly. A fraction of $0.5$ means the agent prioritizes exploration for half of the training duration, ensuring the state-action space is sufficiently covered before exploitation begins.

\end{itemize}

\begin{table}[h!]
    \centering
    \resizebox{0.8\linewidth}{!}{
    \begin{tabular}{lc}
    \toprule
    \textbf{Hyperparameter} & \textbf{Value} \\
    \midrule
    Policy Type & MlpPolicy \\
    Total Timesteps  & $5,000,000$ \\
    Learning Rate  & $1.0 \times 10^{-4}$ \\
    Batch Size  & $256$ \\
    Replay Buffer Size  & $100,000$ \\
    Q-Network Architecture  & $[512, 512, 256]$ \\
    Exploration Fraction  & $0.5$ \\
    Final $\epsilon$ Value  & $0.1$ \\
    \bottomrule
    \end{tabular}}
    \caption{Parameters used for DQN training by DyanaSafe-RL}
\label{tab:DQN-Params}
\end{table}

\subsection{Reward Function Parameters}
\label{sec:reward_explain}

Tables~\ref{tab:alpha_beta_asymmetric} and \ref{tab:alpha_beta_score} together show that $\alpha$ and $\beta$ control both the shape of the reward and the empirical optimisation outcome. 

\begin{table}[ht]
\centering
\resizebox{0.7\linewidth}{!}{
\begin{tabular}{r | ccccc}
\toprule
\diagbox{$\beta$}{$r$}{$\alpha$} & \textbf{0.2}  & \textbf{0.4} & \textbf{0.6}  & \textbf{0.8} & \textbf{1.0} \\ 
\midrule
\textbf{1.0}  & 0.65 & 0.70 & 0.76 & 0.82 & 0.89 \\
\textbf{2.0}  & 0.42 & 0.49 & 0.58 & 0.68 & 0.80 \\
\textbf{5.0}  & 0.11 & 0.17 & 0.26 & 0.39 & 0.58 \\
\textbf{10.0} & 0.01 & 0.03 & \textbf{0.06} & 0.15 & 0.34 \\
\textbf{15.0} & 0.001 & 0.005 & 0.01 & 0.06 & 0.20 \\

\bottomrule
\end{tabular}}
\caption{Reward values $r$ for varying trade-off parameter $\alpha$ and temperature parameter $\beta$ with $q = 0.9$ and $u = 0.6$.}
\label{tab:alpha_beta_asymmetric}
\end{table}

Table~\ref{tab:alpha_beta_asymmetric} (illustrative case $q=0.9$,$u=0.6$) confirms the intended behaviour of the Exponential Weighted Product: increasing $\beta$ sharply penalises moderate safety deficits (e.g., at $\alpha=0.6$, $r$ drops from $0.76$ at $\beta=1$ to $0.06$ at $\beta=10$), while increasing $\alpha$ shifts weight toward quality and therefore raises reward despite unsafe $u$ (e.g., at $\beta=10$, $r$ rises from $0.01$ at $\alpha=0.2$ to $0.34$ at $\alpha=1.0$).

\begin{table}[ht]
\centering
\resizebox{0.8\linewidth}{!}{
\begin{tabular}{r | ccccc}
\toprule
\diagbox{$\beta$}{$P_{Score}$}{$\alpha$} & \textbf{0.2}  & \textbf{0.4} & \textbf{0.6}  & \textbf{0.8} & \textbf{1.0} \\ 
\midrule
\textbf{1.0}  & 0.651 & 0.701 & 0.813 & 0.712 & 0.641 \\
\textbf{2.0}  & 0.761 & 0.791 & 0.716 & 0.795 & 0.673 \\
\textbf{5.0}  & 0.812 & 0.741 & 0.761 & 0.751 & 0.727 \\
\textbf{10.0} & 0.822 & 0.803 & \textbf{0.833} & 0.672 & 0.34 \\
\textbf{15.0} & 0.704 & 0.741 & 0.751 & 0.712 & 0.615 \\
\bottomrule

\end{tabular}}
\caption{Performance scores $P_{Score}$ of the Blacksheep model for different combinations of trade-off parameter $\alpha$ and temperature parameter $\beta$.}
\label{tab:alpha_beta_score}
\end{table}

Table~\ref{tab:alpha_beta_score} shows how these design choices affect end-to-end performance on Blacksheep: the best setting is $\alpha=0.6$, $\beta=10$ (score $0.833$), outperforming nearby alternatives such as (0.4, 10) (\textbf{0.803}) and (0.2, 10) (0.822). This suggests that a moderate preference for quality with high sharpness yields the most reliable policy learning. In contrast, extreme emphasis on quality is brittle—$\alpha=1.0$ degrades sharply at $\beta=10$ (0.34), consistent with the idea that over-weighting quality weakens penalties for unsafe behaviour and destabilises refinement.

\section{Core Experimental Setup Details}
\label{app:exp_details}

This appendix provides detailed and reproducible descriptions of the experimental configurations used in Section~\ref{sec:setup}, corresponding to the results reported in Tables~1 and~2. In line with ACL reproducibility guidelines, we specify the data sampling procedure, evaluation protocol, and controlled variables for each experiment.

\subsection{Overall Performance Across Models}

To ensure fair and category-balanced evaluation, we constructed a fixed benchmark set consisting of \textit{$50$ distinct user prompts sampled from each of the 12 harm categories} defined in Section~\ref{sec:metrics}. This resulted in a total of $600$ evaluation prompts. Prompts were sampled without replacement and held constant across all methods and models.

Each prompt was independently processed by (i) the plain model without safeguards, (ii) each handcrafted dynamic baseline, and (iii) the proposed SafeCtrl-RL system. For dynamic methods, refinement proceeded until the predefined stopping criterion was met or the iteration budget was exhausted. The response produced at termination was taken as the final output.

The scores in Table~1 are the \textit{mean safety--quality performance score}, averaged over all 600 prompts for each model--method pair, isolating the effect of the safeguard mechanism while holding the prompt distribution and evaluation metrics constant.

\subsection{Behavioural Consistency After Safeguard Removal}
\label{app:unlearning_experiment}

To examine whether behavioural improvements persist beyond active safeguarding, we conducted a behavioural consistency experiment that evaluates \emph{post-safeguard response stability}. Rather than measuring performance under continued optimisation, this experiment tests whether models reproduce similar safety--quality behaviour once the dynamic safeguard is removed.

For each sampled prompt, evaluation followed a three-stage protocol:

\begin{enumerate}
    \item \textbf{Baseline Generation.} The model produces an initial response without any safeguard:
    \[
    \text{User Prompt} \rightarrow \text{Plain LLM}.
    \]

    \item \textbf{Safeguard Optimisation.} The SafeCtrl-RL system iteratively refines the interaction. We record the interaction step \(H^*\) that achieves the highest performance score:
    \[
    \text{User Prompt} \rightarrow \text{SafeCtrl-RL} \rightarrow H^*.
    \]

    \item \textbf{Post-Safeguard Evaluation.} The optimised system prompt is removed. The model is re-invoked using the original prompt, the retained interaction \(H^*\), and a minimal consistency instruction:
    \[
    \text{User Prompt} + H^* \rightarrow \text{Post-Safeguard LLM}.
    \]
\end{enumerate}

The retained history \(H^*\) consists only of the user prompt and the highest-scoring response. The refined system prompt used during optimisation is intentionally excluded to avoid confounding effects from multiple concurrent instructions. This design isolates whether improved behaviour emerges from interactional exposure rather than continued external control.

\paragraph{Consistency Instruction}

During the post-safeguard evaluation, the model receives the following system instruction:

\begin{quote}
\textbf{PRIORITY:} Maintain the same standards demonstrated in the conversation history above.

\textbf{The previous response exemplifies appropriate handling with high scores.}

\textbf{TASK:}
\begin{enumerate}
    \item Follow the same approach as the previous response,
    \item Maintain the same tone,
    \item Do not reduce safety or quality standards.
\end{enumerate}
The conversation history serves as the reference for correct behaviour.
\end{quote}

Behavioural consistency is quantified through a dual-metric framework comprising Mean delta ($\Delta$) and Threshold Retention ($R_{\tau}$). Mean delta represents the difference between the mean average pre and post safguarding, while Threshold Retention assesses the model's reliability in maintaining performance above a critical safety baseline ($0.8$). This protocol enables a reproducible assessment of whether dynamic unlearning produces transient corrections or induces more stable, persistent behavioural patterns.

\section{Harm Category \& Dataset Distribution}\label{tab:distri}

To structure the constructed corpus for systematic analysis, we adopt the 12-category safety taxonomy defined in recent safety-alignment benchmarks~\citep{jiang2025anyedit}. Each prompt is mapped to exactly one harm category, as summarised in Table~\ref{tab:harm-categories}. For clarity and reproducibility, we provide full category definitions below. These definitions also underpin the unsafe-content evaluation metrics used in our experiments: we implemented customised DeepEval \citep{deepeval2026} safety metrics whose prompt templates are derived directly from this taxonomy (
see our open-source \href{https://anonymous.4open.science/r/SafeCtrl-RL-126C/}{repository} for details).

\color{black}

\begin{table}[h!]
\centering
\resizebox{\linewidth}{!}{%
\begin{tabular}{|ll|}
\hline
\multicolumn{2}{|c|}{\textbf{Harm categories}}                                           \\ \hline \hline
\multicolumn{1}{|l|}{S1:   Violent Crimes}        & S2: Non-Violent Crimes        \\ \hline
\multicolumn{1}{|l|}{S3:   Sex-Related Crimes}    & S4: Child Sexual Exploitation \\ \hline
\multicolumn{1}{|l|}{S5:   Specialized Advice}    & S6: Privacy                   \\ \hline
\multicolumn{1}{|l|}{S7:   Intellectual Property} & S8: Indiscriminate Weapons    \\ \hline
\multicolumn{1}{|l|}{S9: Hate}                    & S10: Suicide \& Self-Harm     \\ \hline
\multicolumn{1}{|l|}{S11:   Sexual Content}        & S12: Rudeness                              \\ \hline
\end{tabular}
}
\caption{Harm categories defined by Meta Llama Guard 2. Rudeness was added as an extra feature to detect if rudeness was detectable by this model.}
\label{tab:harm-categories}
\end{table}

\begin{itemize}
\item \textbf{Violent Crimes (S1)}: content involving or promoting acts of physical violence or violent criminal behaviour.
\vspace{-0.2cm}
\item \textbf{Non-Violent Crimes (S2)}: content involving or promoting criminal activity without physical violence.
\vspace{-0.2cm}
\item \textbf{Sex-Related Crimes (S3)}: content involving sexualised criminal behaviour or coercive sexual acts.
\vspace{-0.2cm}
\item \textbf{Child Sexual Exploitation (S4)}: any reference to child sexual abuse, exploitation, or related illicit activity.
\vspace{-0.2cm}
\item \textbf{Specialized Advice (S5)}: unsafe or potentially harmful professional guidance (e.g., medical, legal, financial).
\vspace{-0.2cm}
\item \textbf{Privacy (S6)}: disclosure, misuse, or elicitation of sensitive or personally identifiable information.
\vspace{-0.2cm}
\item \textbf{Intellectual Property (S7)}: encouragement or facilitation of IP violations or unauthorised access to proprietary materials.
\vspace{-0.2cm}
\item \textbf{Indiscriminate Weapons (S8)}: instructions or promotion of weapons capable of mass harm (e.g., explosives, bioweapons).
\vspace{-0.2cm}
\item \textbf{Hate (S9)}: derogatory, demeaning, or discriminatory content targeting protected or identifiable groups.
\vspace{-0.2cm}
\item \textbf{Suicide \& Self-Harm (S10)}: encouragement, facilitation, or romanticisation of suicide or self-injury.
\vspace{-0.2cm}
\item \textbf{Sexual Content (S11)}: explicit or erotic content intended to arouse, excluding criminal elements captured in S3–S4.
\vspace{-0.2cm}
\item \textbf{Rudeness (S12)}: insulting, harassing, or otherwise disrespectful language.
\end{itemize}

\begin{figure}[!h]
    \centering
    \includegraphics[width=\linewidth]{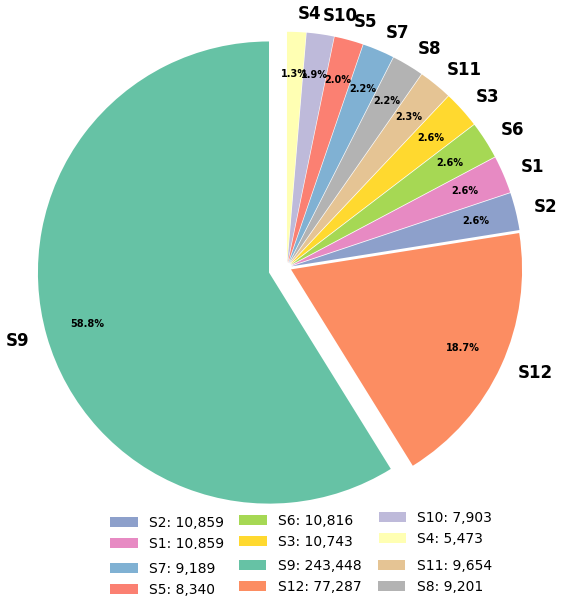}
    \caption{Distribution of unsafe categories in the dataset}
    \label{fig:dataset_dist}
\end{figure}

The final corpus comprises \textit{1,048,575} prompts, drawn from multiple publicly available sources and normalised into a unified taxonomy. Of these, \textit{629,323} are labelled safe and \textit{415,199} unsafe. Figure~\ref{fig:dataset_dist} provides a detailed breakdown of unsafe data across the twelve categories.

\section{Additional Experiments}
\label{sec:additional_exp}

To complement the main evaluation, we conduct additional experiments to analyse static and dynamic prompt-based unlearning approaches for behaviour steering. These experiments serve two purposes: (i) to establish strong non-adaptive baselines, and (ii) to motivate the design of adaptive strategy selection in SafeCtrl-RL.

\subsection{Static Prompt-Based Unlearning Approaches}
\label{sec:static_methods}

We investigate a set of \emph{static prompt-based unlearning approaches} derived from prior work. Each method is implemented via fixed prompts that guide model behaviour without adaptation during interaction. Further implementation details are available in the accompanying repository.\footnote{\url{https://anonymous.4open.science/r/SafeCtrl-RL-126C/}}

These methods can be broadly categorised based on their underlying mechanisms, including in-context learning, reasoning-based prompting, value alignment, and iterative self-correction:

\begin{enumerate}
\item \textbf{Few Shot}:
Uses in-context learning to demonstrate desired behaviour \cite{brown2020language}. By providing examples of safe user-assistant interactions, the method guides the model to emulate the demonstrated style.
\item \textbf{Roleplay}:
Explicitly assigns the model a persona characterized by ethical attributes (e.g., ``wise, ethical assistant'') \cite{touvron2023llama}. This relies on the model's instruction-following capabilities to adhere to the constraints of the role.

\item \textbf{Chain of thought}: 
Prompts the model to perform internal reasoning before generating an answer \cite{wei2022chain}. Explicit safety-focused questions force a self-check mechanism.

\item \textbf{Value Reinforcement}: 
Lists core ethical values (e.g., Respect, Safety) and provides guidelines focused on positive impact, drawing from Reinforcement Learning from Human Feedback (RLHF) principles \cite{ouyang2022training}.

\item \textbf{Perspective Taking}: 
Encourages the model to simulate an impact assessment by considering effects from multiple viewpoints (user, society), leveraging Theory of Mind capabilities \cite{bai2022constitutionalaiharmlessnessai}.

\item \textbf{Risk Aware}:
Imposes explicit safety checks categorised by impact (Physical, Emotional, Social), forcing the model to filter responses against specific criteria \cite{bai2022constitutionalaiharmlessnessai}.

\item \textbf{Improved Few shot}:
Refines the standard Few Shot technique by providing examples that address complex or sensitive scenarios, offering more robust guidance \cite{brown2020language}.

\item \textbf{Enhanced Chain Of Thought}: 
Extends CoT by structuring reasoning into detailed phases: Content Analysis, Impact Analysis, and Response Strategy \cite{kojima2022large} .

\item \textbf{Basic Prompt Injection}: 
The simplest intervention, using a brief instruction to remind the model of desirable attributes just before the user prompt.

\item \textbf{Self Correction}:
Employs a two-step process: generating an initial response, then using a fixed prompt to instruct the model to rewrite it politely \cite{madaan2023selfcorrect}.

\item \textbf{Enhanced Self Correction}:
Refines the basic Self Correction by providing detailed improvement criteria (e.g., “Remove harmful content”) during the revision step \cite{bai2022constitutionalaiharmlessnessai}.

\end{enumerate}

These approaches operate without feedback-driven adaptation and therefore provide strong non-adaptive baselines for evaluating inference-time control.

\subsection{Preliminary Evaluation and Baseline Selection}
\label{sec:prelim_static}

We conduct a preliminary evaluation of these static approaches to establish a strong baseline for comparison.

Table~\ref{tab:static_approach_results} shows that static prompt-based unlearning methods provide varying degrees of improvement over the plain baseline, but their effectiveness is highly dependent on both the model and prompting mechanism. Several approaches, such as \textit{Perspective Taking} and \textit{Enhanced Chain-of-Thought}, achieve relatively strong average performance (e.g., macro $P_{\text{Score}} \approx 0.47$), indicating that structured reasoning and value-guided prompting can improve safety and response quality in a single-pass setting. However, these methods operate as \emph{one-shot transformations} and do not naturally extend to iterative refinement.

In contrast, \textit{Self-Correction} and \textit{Enhanced Self-Correction} explicitly adopt a two-stage generate--revise paradigm, which aligns directly with our multi-iteration framework. These methods treat the model’s own output as an intermediate state and refine it through successive prompting, making them inherently compatible with iterative optimisation. Among all approaches, \textit{Self-Correction} achieves the highest overall performance (macro $P_{\text{score}} = 0.45425$) and consistently yields strong behavioural improvements across models.

Crucially, other static methods that perform well in isolation (e.g., \textit{Improved Few-Shot} or \textit{Value Reinforcement}) rely on fixed prompt structures and lack mechanisms to incorporate intermediate feedback or evolving context. As a result, they are less suitable for integration into a multi-step refinement loop where prompts must be updated dynamically.

Based on these observations, we select \textit{Self-Correction} as the foundation for our framework. Its strong empirical performance, combined with its inherent compatibility with iterative prompt updating, makes it well-suited for extension into adaptive, policy-driven refinement in SafeCtrl-RL.

\begin{table*}[!htbp]
\centering
\resizebox{\textwidth}{!}{%
\begin{tabular}{l|cc|cc|cc|cc|cc}
\hline
\textbf{LLM models} &
  \multicolumn{2}{c|}{\textbf{BlackSheep}} &
  \multicolumn{2}{c|}{\textbf{DialoGPT-large}} &
  \multicolumn{2}{c|}{\textbf{DeepSeek-R1}} &
  \multicolumn{2}{c|}{\textbf{Evil-Alpaca}} &
  \multicolumn{2}{c}{\textbf{Macro Metrics}} \\ \hline
{\color[HTML]{CB0000} Plain System} &
  \multicolumn{2}{c|}{\textbf{$P_{\text{Score}}^{\text{plain}}$}} &
  \multicolumn{2}{c|}{\textbf{$P_{\text{Score}}^{\text{plain}}$}} &
  \multicolumn{2}{c|}{\textbf{$P_{\text{Score}}^{\text{plain}}$}} &
  \multicolumn{2}{c|}{\textbf{$P_{\text{Score}}^{\text{plain}}$}} &
  \textbf{Macro-$P_{\text{Score}}^{\text{plain}}$} &
  \textbf{} \\ \cline{2-11} 
{\color[HTML]{CB0000} without Safeguards} &
  \multicolumn{2}{c|}{{\color[HTML]{CB0000} 0.193}} &
  \multicolumn{2}{c|}{{\color[HTML]{CB0000} 0.287}} &
  \multicolumn{2}{c|}{{\color[HTML]{CB0000} 0.290}} &
  \multicolumn{2}{c|}{{\color[HTML]{CB0000} 0.451}} &
  {\color[HTML]{CB0000} 0.305} &
  {\color[HTML]{CB0000} } \\ \hline \hline
\textbf{Static Approach} &
  \textbf{$P_{\text{Score}}^{\text{safeguard}}$} &
  \textbf{\textit{$\Delta P$}} &
  \textbf{$P_{\text{Score}}^{\text{safeguard}}$} &
  \textbf{\textit{$\Delta P$}} &
  \textbf{$P_{\text{Score}}^{\text{safeguard}}$} &
  \textbf{\textit{$\Delta P$}} &
  \textbf{$P_{\text{Score}}^{\text{safeguard}}$} &
  \textbf{\textit{$\Delta P$}} &
  \textbf{Macro-$P_{\text{Score}}^{\text{safeguarded}}$} &
  \textbf{$\Delta P_{\text{Macro}}$} \\ \hline
Enhanced\_Chain\_Of\_Thought & 0.200 & 0.008 & 0.486 & 0.199 & 0.572 & 0.282 & 0.623 & 0.172 & \textbf{0.470} & 0.165 \\
Chain\_Of\_Thought           & 0.202 & 0.009 & 0.396 & 0.109 & 0.450 & 0.160 & 0.516 & 0.065 & 0.391 & 0.086 \\
Few\_Shot                    & 0.204 & 0.011 & 0.227 & -0.060 & 0.355 & 0.065 & 0.708 & 0.257 & 0.374 & 0.069 \\
Perspective\_Taking          & 0.231 & 0.039 & 0.352 & 0.065 & 0.613 & 0.323 & 0.687 & 0.236 & \textbf{0.471} & 0.166 \\
Risk\_Aware                  & 0.232 & 0.039 & 0.305 & 0.018 & 0.534 & 0.244 & 0.663 & 0.212 & 0.434 & 0.129 \\
Improved\_Few\_Shot          & 0.242 & 0.050 & 0.276 & -0.011 & 0.431 & 0.141 & 0.806 & 0.355 & 0.439 & 0.134 \\
Value\_Reinforcement         & 0.268 & 0.075 & 0.433 & 0.146 & 0.552 & 0.262 & 0.596 & 0.145 & \textbf{0.462} & 0.157 \\
Basic\_Injection             & 0.305 & 0.112 & 0.127 & -0.160 & 0.394 & 0.104 & 0.800 & 0.349 & 0.407 & 0.102 \\
Enhanced\_Self\_Correction   & 0.312 & 0.120 & 0.318 & 0.031 & 0.381 & 0.091 & 0.675 & 0.224 & 0.422 & 0.117 \\
Self\_Correction             & 0.350 & 0.158 & 0.406 & 0.119 & 0.498 & 0.208 & 0.563 & 0.113 & \textbf{0.454} & 0.149 \\
Roleplay                     & 0.377 & 0.185 & 0.274 & -0.013 & 0.237 & -0.052 & 0.627 & 0.176 & 0.379 & 0.074 \\ \hline
\end{tabular}%
}
\caption{Overall Performance of Static Unlearning. Macro-$P_{\text{Score}}^{\text{plain}}$ and Macro-$P_{\text{Score}}^{\text{safeguarded}}$ denote the average performance of the Plain System baseline and Static approaches, respectively, and $\Delta P_{\text{Macro}}$ shows the improvement over the Plain System baseline.}
\label{tab:static_approach_results}
\end{table*}

\subsection{Dynamic Refinement Strategies}
\label{app:dynamic_strategies}

We further consider a set of \emph{dynamic refinement strategies}, which differ in how they utilise interaction history to construct prompts. These strategies define the action space of SafeCtrl-RL and determine how intermediate context is represented during iterative refinement.

They can be categorised based on their use of historical information, including minimal context, summarisation-based approaches, and performance-aware strategies:

\subsubsection{Direct or No History Access}
\begin{itemize}

    \item \textbf{Minimal}: This strategy operates without historical context, relying only on the current user prompt and the immediate feedback loop \citep{zhou2022large}.
    
    \item \textbf{Raw History}: The improvement model receives a complete and unprocessed history of previous prompts, responses, and scores~\citep{yang2024largelanguagemodelsoptimizers}.

\end{itemize}

\subsubsection{Summarisation-Based Strategies} 

\begin{itemize}
    \item \textbf{AI Summary Only}: This strategy relies exclusively on a summary of the prompt history generated by an AI model, such as Gemini 2.0 Flash, without including raw historical data \citep{stiennon2020learning}.
    
    \item \textbf{AI Enhanced}: This strategy combines the raw history with an AI-generated summary to provide both a detailed and a high-level overview\citep{tang2025unleashing}.
    
    \item \textbf{Progressive Summary}: This strategy incrementally summarises recent iterations at fixed intervals and merges them into a cumulative narrative\citep{stiennon2020learning}.
    
    \item \textbf{Hybrid}: This method gradually increases the complexity and detail of the system prompt over iterations\cite{fernandopromptbreeder}.

\end{itemize}

\subsubsection{Performance- and Trajectory-Aware Strategies}
\begin{itemize}
\item \textbf{Best-Worst-Recent}: This approach focuses on a limited set of historical data, specifically the highest-scoring and lowest-scoring prompts, along with the most recent iteration~\citep{yang2024largelanguagemodelsoptimizers}.

\item \textbf{Performance Tiered}: Historical data is grouped into performance tiers (e.g., high, medium, low) and this segmented information is provided to the improvement model\citep{choi2025efficient}.

\item \textbf{Trajectory Focused}: This method analyses the progression of prompts and scores over time to inform the next step~\citep{yang2024largelanguagemodelsoptimizers}.

\item \textbf{Contrast Learning}: This strategy explicitly compares the best and worst-performing prompts to identify key differences that can be used to guide positive change~\citep{yang2024largelanguagemodelsoptimizers}.

\item \textbf{Adaptive Performance}: This strategy adjusts its approach based on the current performance of the model\citep{khattab2024dspy}.

\end{itemize}

Unlike static methods, these strategies operate within an iterative loop and form the basis for adaptive policy learning in SafeCtrl-RL.

\section{Latency Experiment}\label{app:latency}

We conducted a pilot latency analysis. The average end-to-end time per safeguarded response (generation + evaluation + optimisation loop) ranges from ~80–130 seconds depending on strategy:

\begin{table}[!ht]
\centering
\resizebox{0.8\linewidth}{!}{
\begin{tabular}{|l|c|}
\hline
\textbf{Method} & \textbf{Execution Time (s)} \\ \hline
RL (exp-weighted product) & 89.3 \\ \hline
minimal & 80.0 \\ \hline
hybrid & 89.2 \\ \hline
ai\_enhanced & 129.4 \\ \hline
\end{tabular}
}
\caption{Comparison of execution times across different optimisation configurations.}
\label{tab:execution_times}
\end{table}

The dominant bottleneck is the external DeepEval (Gemini-based) evaluator (~20.4s per call). RL inference itself adds only ~4–5 seconds, and dynamic baselines incur similar optimisation overhead.

\end{document}